\documentclass[twocolumn]{svjour3}          
\smartqed  
\usepackage{graphicx}
\usepackage{mathptmx}      
\usepackage{times}
\usepackage{epsfig}
\usepackage{graphicx}
\usepackage{amsmath}
\usepackage{amssymb}
\usepackage{caption}
\usepackage[pagebackref=true,breaklinks=true,letterpaper=true,colorlinks,bookmarks=false]{hyperref}
\usepackage{enumerate}
\usepackage{booktabs} 
\usepackage{array}  %
\usepackage{multirow}
\usepackage[tight,normalsize,sf,SF]{subfigure}
\usepackage{animate}
\usepackage{bbding}
\usepackage{blindtext}
\newcommand{\etal}{\textit{et al}.}
\newcommand{\ie}{\textit{i}.\textit{e}.}
\newcommand{\eg}{\textit{e}.\textit{g}.}
\newcommand{\etc}{\textit{etc}.}

\makeatletter
\DeclareRobustCommand\onedot{\futurelet\@let@token\@onedot}
\def\@onedot{\ifx\@let@token.\else.\null\fi\xspace}
\def\vs{\emph{vs}\onedot}
\makeatother

\def\myTextColor{\textcolor[rgb]{0, 0, 0}}
\journalname{Noname}
\begin{document}
\begin{sloppypar}

\title{Occluded Video Instance Segmentation: A Benchmark 
}


\author{Jiyang~Qi~\textsuperscript{1,2}\textsuperscript{*} \and Yan~Gao~\textsuperscript{2}\textsuperscript{*} \and Yao~Hu~\textsuperscript{2} \and Xinggang~Wang~\textsuperscript{1} \and Xiaoyu~Liu~\textsuperscript{2} \and Xiang~Bai~\textsuperscript{1} \and Serge~Belongie~\textsuperscript{3} \and Alan~Yuille~\textsuperscript{4} \and Philip~H.S.~Torr~\textsuperscript{5} \and Song~Bai~\textsuperscript{2,5}~\Envelope}

\authorrunning{Jiyang Qi\textsuperscript{*}, Yan Gao\textsuperscript{*}~\etal} 

\institute{
Song Bai~~\textsuperscript{2,5}~\Envelope \at
\email{songbai.site@gmail.com}
\and
Jiyang Qi~\textsuperscript{1,2}~\footnotemark[1] \at
\email{jiyangqi@hust.edu.cn}
\and
Yan Gao~\textsuperscript{2}~\footnotemark[1] \at
\email{yangao0119@gmail.com}
\and
Yao Hu~\textsuperscript{2} \at
\email{yaoohu@alibaba-inc.com}
\and
Xinggang Wang~\textsuperscript{1} \at
\email{xgwang@hust.edu.cn}
\and
Xiaoyu Liu~\textsuperscript{2} \at
\email{xiaoyuliu1991xyl@gmail.com}
\and
Xiang Bai~\textsuperscript{1} \at
\email{xbai@hust.edu.cn}
\and
Serge Belongie~\textsuperscript{3} \at
\email{s.belongie@di.ku.dk}
\and
Alan Yuille~\textsuperscript{4} \at
\email{alan.l.yuille@gmail.com}
\and
Philip H.S. Torr~\textsuperscript{5} \at
\email{philip.torr@eng.ox.ac.uk}
\and
\textsuperscript{1} Huazhong University of Science and Technology, Wuhan, China \at
\textsuperscript{2} Alibaba Group, Beijing, China \\
\textsuperscript{3} University of Copenhagen, Copenhagen, Denmark \\
\textsuperscript{4} Johns Hopkins University, Baltimore, USA \\
\textsuperscript{5} University of Oxford, Oxford, UK \\
\footnotemark[1] indicates equal contributions. \\
\Envelope~~Corresponding Author
}

\date{Received: date / Accepted: date}

\maketitle

\begin{abstract}

Can our video understanding systems perceive objects when a heavy occlusion exists in a scene?

To answer this question, we collect a large-scale dataset called OVIS for occluded video instance segmentation, that is, to simultaneously detect, segment, and track instances in occluded scenes. OVIS consists of 296k high-quality instance masks from 25 semantic categories, where object occlusions usually occur. While our human vision systems can understand those occluded instances by contextual reasoning and association, our experiments suggest that current video understanding systems cannot. On the OVIS dataset, the highest AP achieved by state-of-the-art algorithms is only 16.3, which reveals that we are still at a nascent stage for understanding objects, instances, and videos in a real-world scenario.
We also present a simple plug-and-play module that performs temporal feature calibration to complement missing object cues caused by occlusion. Built upon MaskTrack R-CNN and SipMask, we obtain a remarkable AP improvement on the OVIS dataset.
The OVIS dataset and project code are available at \href{http://songbai.site/ovis}{http://songbai.site/ovis}.



\end{abstract}

\section{Introduction}

In the visual world, objects rarely occur in isolation. The psychophysical and computational studies~\cite{nakayama1989stereoscopic,hegde2008preferential} have demonstrated that human vision systems can perceive heavily occluded objects with contextual reasoning and association. The question then becomes,~\textit{can our video understanding system perceive objects that are severely obscured?}

\begin{figure}[tb]
\centering
   \subfigure
   {
      \animategraphics[width=0.45\linewidth,height=0.253\linewidth, autopause, poster=4]{3}{imgs-fig1_anime-2930398-}{0}{6}
   }
  \vspace{-2mm}
   \subfigure
   {
      \animategraphics[width=0.45\linewidth,height=0.253\linewidth, autopause, poster=2]{3}{imgs-fig1_anime-3021160-}{0}{5}
   }
  \vspace{-2mm}
   \subfigure
   {
      \animategraphics[width=0.45\linewidth,height=0.253\linewidth, autopause, poster=9]{3}{imgs-fig1_anime-2592056-}{0}{9}
   }
   \subfigure
   {
      \animategraphics[width=0.45\linewidth,height=0.253\linewidth, autopause, poster=8]{3}{imgs-fig1_anime-2932104-}{0}{8}
   }
   \subfigure
   {
      \animategraphics[width=0.45\linewidth,height=0.253\linewidth, autopause, poster=2]{3}{imgs-fig1_anime-2524877_0_170-}{0}{5}
   }
   \subfigure
   {
      \animategraphics[width=0.45\linewidth,height=0.253\linewidth, autopause, poster=first]{3}{imgs-fig1_anime-2932109-}{0}{6}
   }
   \caption{Sample video clips from OVIS. \textcolor{red}{\textbf{Click}} them to watch the animations (best viewed with Acrobat/Foxit Reader). The hairs and whiskers of animals are all exhaustively annotated.}
   \label{fig:anime}
\end{figure}

Our work aims to explore this matter in the context of video instance segmentation, a popular task proposed in~\cite{youtube_vis} that targets a comprehensive understanding of objects in videos. To this end, we explore a new and challenging scenario called \textbf{O}ccluded \textbf{V}ideo \textbf{I}nstance \textbf{S}egmentation (\textbf{OVIS}), which requests a model to simultaneously detect, segment, and track object instances in occluded scenes.

\begin{figure*}[t]
\centering
\includegraphics[width=0.98\linewidth]{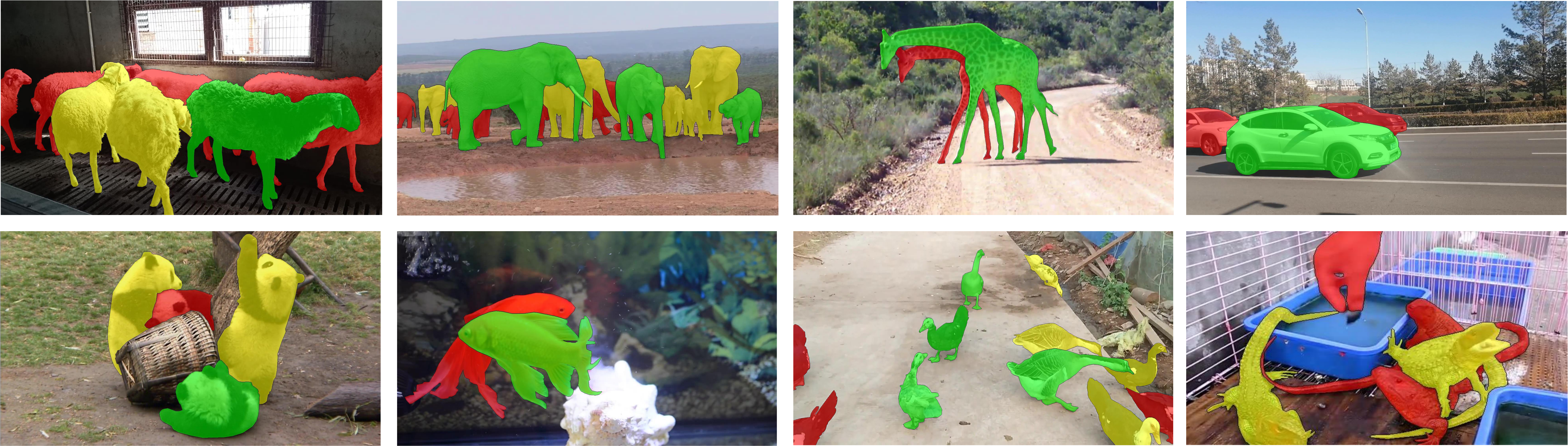}
\caption{Different occlusions levels in OVIS. Unoccluded objects are colored green, slightly occluded objects are colored yellow, and severely occluded objects are colored red.}
\label{fig:various_occlusion}
\end{figure*}

As the major contribution of this work, we collect a large-scale dataset called OVIS, specifically for video instance segmentation in occluded scenes. While being the second video instance segmentation dataset after YouTube-VIS~\cite{youtube_vis}, OVIS consists of 296k high-quality instance masks out of 25 commonly seen semantic categories. Some example clips are given in Fig.~\ref{fig:anime}. The most distinctive property of OVIS dataset is that most objects are under severe occlusions. The occlusion level of each object is also labeled (as shown in Fig.~\ref{fig:various_occlusion}) and we also present an AP (average precision) based metric to measure performance under different occlusion degrees. Therefore, OVIS is a useful testbed to evaluate video instance segmentation models for dealing with heavy object occlusions.

To dissect the OVIS dataset, we conduct a thorough evaluation of 9 state-of-the-art algorithms whose code is publicly available, including FEELVOS~\cite{feelvos}, IoUTracker+~\cite{youtube_vis},  MaskTrack R-CNN~\cite{youtube_vis}, SipMask~\cite{sipmask}, STEm-Seg~\cite{stem_seg}, STMask~\cite{stmask}, TraDeS~\cite{trades}, CrossVIS~\cite{crossvis}, and QueryVIS~\cite{queryinst}. However, the experimental results suggest that current video understanding systems fall behind the capability of human beings in terms of occlusion perception. The highest AP is only 16.3 achieved by~\cite{crossvis} and the highest AP on the heavily occluded group is only 6.3 achieved by~\cite{stmask}. In this sense, we are still far from deploying those techniques into practical applications, especially considering the complexity and diversity of scenes in the real visual world.

To alleviate the occlusion issue, we also present a plug-and-play module called temporal feature calibration. For a given query frame in a video, we resort to a reference frame to complement its missing object cues. Specifically, the proposed module learns a calibration offset for the reference frame with the guidance of the query frame, and then the offset is used to adjust the feature embedding of the reference frame via deformable convolution~\cite{deformable}. The refined reference embedding is used in turn to assist the object recognition of the query frame. Our module is a highly flexible plug-in. While applied to MaskTrack R-CNN~\cite{youtube_vis} and SipMask~\cite{sipmask} respectively, we obtain an AP of 15.4 and 14.3, significantly outperforming the corresponding baselines by 4.6 and 4.1 in AP respectively.

To summarize, our contributions are three-fold:
\begin{itemize}
\item We advance occlusion handling and video instance segmentation by releasing a new benchmark dataset named \textbf{OVIS} (short for \textbf{O}ccluded \textbf{V}ideo \textbf{I}nstance \textbf{S}egmentation). OVIS is designed with the philosophy of perceiving object occlusions in videos, which could reveal the complexity and the diversity of real-world scenes.
\item We streamline the research over the OVIS dataset by conducting a comprehensive evaluation of 9 state-of-the-art video instance segmentation algorithms, which could be a baseline reference for future research on OVIS.
\item \myTextColor{As a minor contribution, we present a plug-and-play module called Temporal Feature Calibration to alleviate the occlusion issue. Using MaskTrack R-CNN~\cite{youtube_vis} and SipMask~\cite{sipmask} as baselines, the proposed module obtains remarkable improvements on both OVIS and YouTube-VIS. More importantly, its ``plug-and-play" nature makes it widely applicable to future endeavors on OVIS.}
\end{itemize}

\myTextColor{Compared with our conference version~\cite{ovis_nips} that briefly describes the OVIS dataset and challenge held in 2021, the improvements are concluded as follows: 1) more thorough experiments (\eg, oracle experiments, error analysis, per-class result analysis) are conducted to dissect the OVIS dataset and the occlusion problem; 2) we comprehensively evaluate the effect of leveraging temporal context and adjusting the NMS threshold adaptively on occlusion handling; 3) more baseline results (\eg, the results that training with augmented image sequences, the results obtained with larger backbone or larger input resolutions) are provided, which can be a better reference for future work; 4) we further summarize remaining difficulties and future directions that deserve attention in OVIS.}

\section{Related Work}

\subsection{Video Instance Segmentation}

Our work focuses on Video Instance Segmentation in occluded scenes. The most relevant work to ours is~\cite{youtube_vis}, which formally defines the concept of video instance segmentation and releases the first dataset called YouTube-VIS. Built upon the large-scale video object segmentation dataset YouTube-VOS~\cite{youtube_vos}, the 2019 version of YouTube-VIS dataset contains a total of 2,883 videos, 4,883 instances, and 131k masks in 40 categories. Its latest 2021 version contains a total of 3,859 videos, 8,171 instances, and 232k masks. While YouTube-VIS is not designed to study the occluded video understanding problem, most objects in the OVIS dataset are under severe occlusions. Our experimental results show that OVIS is much more challenging.

Since the release of the YouTube-VIS dataset, video instance segmentation has attracted great attention in the computer vision community, arising a series of algorithms recently. MaskTrack R-CNN~\cite{youtube_vis} is the first unified model for video instance segmentation. It fulfills video instance segmentation by adding a tracking branch to the popular image instance segmentation method Mask R-CNN~\cite{maskrcnn}. Lin \textit{et al.}~\cite{vist_vae} propose a modified variational auto-encoder architecture built on the top of Mask R-CNN. 
MaskProp~\cite{maskprop} is also a video extension of Mask R-CNN which adds a mask propagation branch to track instances by the propagated masks. SipMask~\cite{sipmask} extends single-stage image instance segmentation to the video level by adding a fully-convolutional branch for tracking instances. STMask~\cite{stmask} improves feature representation by spatial feature calibration and temporal feature fusion. Different from those top-down methods, STEm-Seg~\cite{stem_seg} proposes a bottom-up method, which performs video instance segmentation by clustering the pixels of the same instance. Built upon Transformers, VisTR~\cite{vistr} supervises and segments instances at the sequence level as a whole. IFC~\cite{ifc} further reduces the computations of full space-time transformers by only executing attention between memory tokens. QueryVIS~\cite{queryinst} follows a multi-stage paradigm and leverages the intrinsic one-to-one correspondence in queries across different stages. Based on FCOS~\cite{fcos}, SGNet~\cite{sgnet} dynamically divides instances into sub-regions and performs segmentation on each region. CrossVIS~\cite{crossvis} uses the instance feature in the current frame to localize the same instance in other frames. Different from the tracking-by-detection paradigm, TraDeS~\cite{trades} integrates tracking cues to assist detection.

\subsection{Other Related Tasks}

Our work is also relevant to several other tasks, including:

\paragraph{Video Object Segmentation.}~Video object segmentation (VOS) is a popular task in video analysis. According to whether to provide the mask for the first frame, VOS can be divided into semi-supervised and unsupervised scenarios. Semi-supervised VOS~\cite{onlinevos_1,onlinevos_2,onlinevos_3,vos1,vos2,vos3,stm,li2020delving,swiftnet} aims to track and segment a given object with a mask. Many Semi-supervised VOS methods~\cite{onlinevos_1,onlinevos_2,onlinevos_3} adopt an online learning manner which fine-tunes the network on the mask of the first frame during inference. Recently, some other works~\cite{vos1,vos2,vos3,stm,li2020delving,swiftnet} aim to avoid online learning for the sake of faster inference speed. Unsupervised VOS methods~\cite{uvos1,uvos2,uvos3} aim to segment the primary objects in a video without the first frame annotations.

As the first video object segmentation dataset, DAVIS~\cite{davis2016,davis2017_unsupervised} contains 150 videos and 376 densely annotated objects. \cite{youtube_vos} further proposes the larger YouTube-VOS dataset with 4,453 video clips and 7,755 objects based on the large-scale YouTube-8M~\cite{youtube-8m} dataset.
Different from video instance segmentation that needs to classify objects, both unsupervised and semi-supervised VOS does not distinguish semantic categories. In addition, only one or several salient objects are annotated in these VOS datasets, while we annotate all the objects belonging to the pre-defined category set.

\paragraph{Video Semantic Segmentation.}~Video semantic segmentation requires semantic segmentation for each frame in a video. The popular video semantic segmentation datasets include Cityscapes~\cite{cityscapes}, CamVid~\cite{camvid},~\etc~There are 5000 video clips in the Cityscapes~\cite{cityscapes} dataset. Each clip consists of 30 frames and only the 20th frame is annotated. CamVid~\cite{camvid} dataset contains 4 videos and the authors annotate one frame every 30 frames, obtaining 800 annotated frames finally.
LSTM~\cite{vss_lstm}, GRU~\cite{vss_gru}, and optical flow~\cite{deep_feature_flow} are introduced to leverage temporal contextual information for more accurate or faster video semantic segmentation. Video semantic segmentation does not require distinguishing instances and tracking objects across frames.

\paragraph{Video Panoptic Segmentation.}~Dahun~\etal~\cite{kim2020video} define a video extension of panoptic segmentation~\cite{panoptic0}, which requires generating consistent panoptic segmentation, and in the meantime, associating instances across frames. They further reformatted the VIPER dataset with 124 videos and proposed the Cityscapes-VPS dataset which contains 500 videos.

\paragraph{Open-World Video Object Segmentation.}~Different from the aforementioned tasks, open-world video object segmentation~\cite{uvo} is taxonomy-free and requires segmenting and tracking all the objects class-agnostically. The proposed UVO dataset~\cite{uvo} contains 1200 videos and all the videos are densely annotated.

\myTextColor{\paragraph{Multi-Object Tracking.}~Multi-object tracking (MOT)~\cite{mot} aims to detect the bounding boxes of objects and track them in a given video. Some popular datasets focus on the tracking of pedestrians and cars in street scenes, such as MOT16~\cite{mot16} and KITTI~\cite{kitti}. Meanwhile, UA-DETRA~\cite{ua_detrac} features vehicle tracking only.}

\paragraph{Multi-Object Tracking and Segmentation.}~Multi-object tracking and segmentation (MOTS)~\cite{mots} extends multi-object tracking (MOT)~\cite{mot} from a bounding box level to a pixel level.
Paul \textit{et al.}~\cite{mots} release the KITTI MOTS and MOTSChallenge datasets, and propose Track R-CNN that extends Mask R-CNN by 3D convolutions to incorporate temporal context and an extra tracking branch for object tracking. Xu~\textit{et al.}~\cite{apolo-mots} release the ApolloScape dataset which provides more crowded scenes and proposes a new track-by-points paradigm. \myTextColor{The task definition of MOTS is similar to video instance segmentation, which means an algorithm needs to simultaneously detect, segment, and track objects. While MOTS mainly focuses on pedestrians and cars in the streets, VIS targets more diverse scenes and more general objects in our daily life, such as animals.}

\myTextColor{\paragraph{Video Object Detection.}~Video object detection (VOD) is a direct extension of image-level object detection. Compared with multi-object tracking, the video object detection task does not require tracking an object. Some commonly used datasets include the ImageNet-VID dataset~\cite{imagenet}, which contains 3862 snippets for training, 555 snippets for validation, and 937 snippets for evaluation.}

Our work is of course relevant to some image-level recognition tasks, such as semantic segmentation~\cite{fcn,deeplapv1,deeplabv3}, instance segmentation~\cite{maskrcnn,maskscore,pointrend}, panoptic segmentation~\cite{panoptic0,panoptic1,panoptic2}, large vocabulary instance segmentation~\cite{lvis,lvis1},~\etc

\subsection{Occlusion Understanding}

There are also some works focusing on occlusion understanding and handling. BCNet~\cite{bcnet} adds a new branch to infer the occluders and utilizes the obtained occluder features to enhance the feature of occludees. OCFusion~\cite{lazarow2020learning} introduces the occlusion head to indicate the occlusion relation between each pair of mask proposals. \cite{deocclusion} proposes a self-supervised method that can recover the occlusion ordering and complete the invisible parts of occluded objects. Different from the full-DNN paradigm described above, Some methods\cite{compositional,compositional2,compositional3} integrate compositional models and deep convolutional neural networks into a unified model which is more robust to partial occlusions. As for pedestrian detection in crowded scenes, \cite{repulsion,orcnn} propose new loss functions to enforce predicted boxes to locate compactly to the corresponding ground-truth objects while far from other objects. \cite{bibox}  regresses two bounding boxes for each object to localize the full body and visible part of a pedestrian respectively. \cite{adaptivenms} introduces adaptive-NMS which adaptively increases the NMS threshold in crowd scenes. \cite{tced} aggregates the temporal context to enhance the feature representations. \cite{oneproposal} predicts multiple instances in one proposal. \myTextColor{In Multi-Object Tracking, \cite{mot_attn_occ1,mot_attn_occ2} utilize the attention module to attend to the visible parts of objects. \cite{mot_topo_occ1,mot_topo_occ2} exploit the topology between objects to track the occluded objects.}
In our experiments, to test the effect of temporal aggregation on occlusion handling, a temporal feature calibration module is presented, in which the calibrated features from neighboring frames are fused with the current frame for reasoning occluded objects and improving the recognition in each frame.

\section{OVIS Dataset}

Given an input video, video instance segmentation requires detecting, segmenting, and tracking object instances simultaneously from a predefined set of object categories. An algorithm is supposed to output the class label, confidence score, and a sequence of binary masks of each instance.

The focus of this work is on collecting a large-scale benchmark dataset for video instance segmentation with severe object occlusions. In this section, we mainly review the data collection process, the annotation process, and the dataset statistics.

\begin{figure}[t]
\centering
         \includegraphics[width=0.95\linewidth]{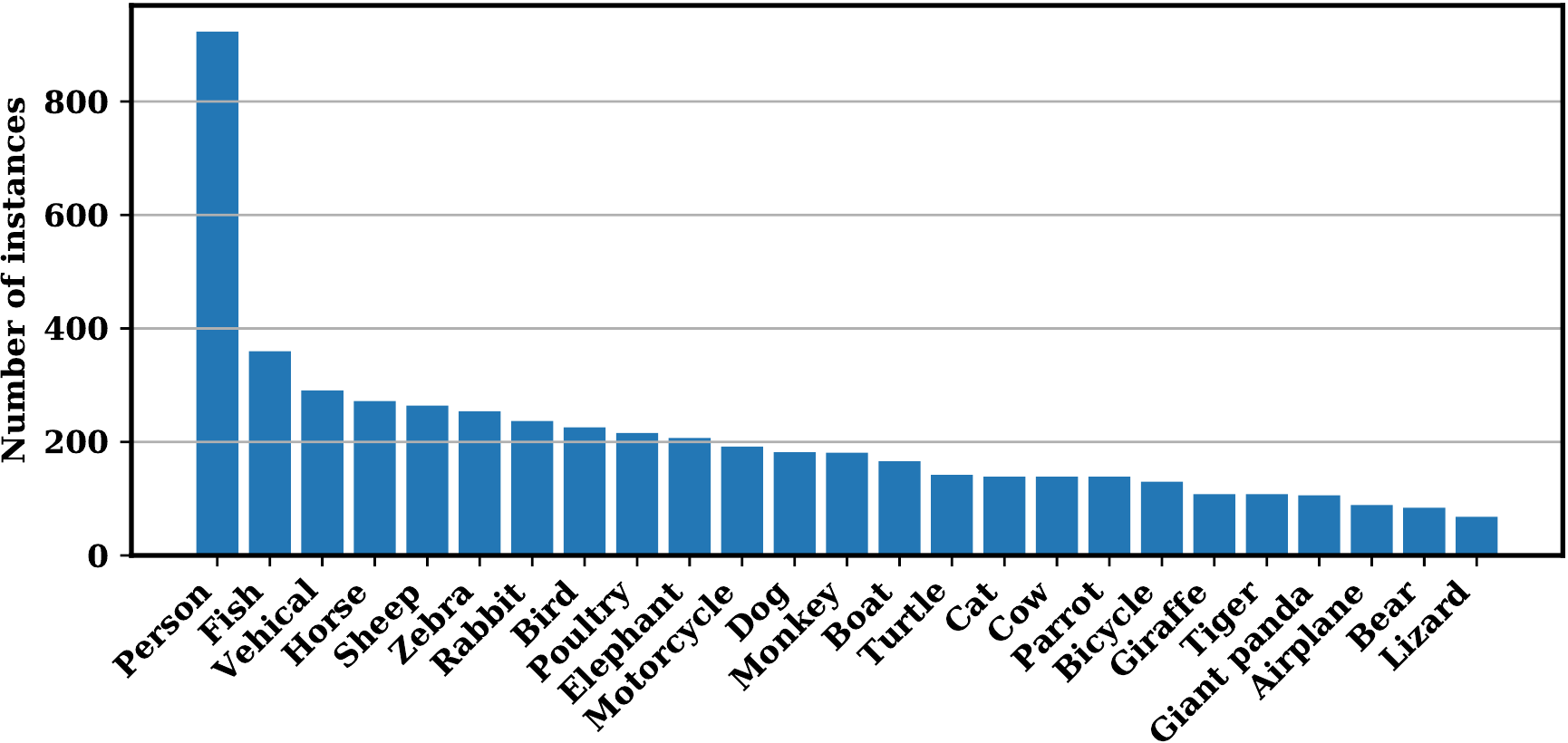}
         \caption{Number of instances per category in the
         OVIS dataset.}
      \label{fig:instance_per_cate}
\end{figure}

\subsection{Video Collection}

We begin with selecting 25 semantic categories, including \textit{Person}, \textit{Bird}, \textit{Cat}, \textit{Dog}, \textit{Horse}, \textit{Sheep}, \textit{Cow}, \textit{Elephant}, \textit{Bear}, \textit{Zebra}, \textit{Giraffe}, \textit{Poultry}, \textit{Giant panda}, \textit{Lizard}, \textit{Parrot}, \textit{Monkey}, \textit{Rabbit}, \textit{Tiger}, \textit{Fish}, \textit{Turtle}, \textit{Bicycle}, \textit{Motorcycle}, \textit{Airplane}, \textit{Boat}, and \textit{Vehicle}. The categories are carefully chosen mainly for three motivations: 1) most of them are animals, with which object occlusions extensively happen, 2) they are commonly seen in our life, 3) these categories have a high overlap with popular large-scale image instance segmentation datasets~\cite{coco,lvis} so that models trained on those datasets are easier to be transferred. The number of instances per category is given in Fig.~\ref{fig:instance_per_cate}.

As the dataset is to study the capability of our video understanding systems to perceive occlusions, we ask the annotation team to 1) exclude those videos, where only one single object stands in the foreground; 2) exclude those videos with a clean background; 3) exclude those videos, where the complete contour of objects is visible all the time. Some other objective rules include 1) video length is generally between 5s and 60s, and 2) video resolution is generally $1920\times1080$.

After applying the objective rules, the annotation team delivers 8,644 video candidates and our research team only accepts 901 challenging videos after a careful re-check. It should be mentioned that due to the stringent standard of video collection, the pass rate is as low as 10\%. 

\subsection{Annotation}
\label{annotation}

Given an accepted video, the annotation team is asked to exhaustively annotate all the objects belonging to the pre-defined category set. Each object is given an instance identity and a class label. In addition to some common rules (\eg,~no ID switch, mask fitness $\leq$1 pixel), the annotation team is trained with several criteria particularly about occlusions: 1) if an existing object disappears because of full occlusions and then re-appears, the instance identity should keep the same; 2) if a new instance appears in an in-between frame, a new instance identity is needed; and 3) the case of ``object re-appears" and ``new instances" should be distinguishable by you after you watch the contextual frames therein. All the videos are annotated every 5 frames and the final annotation granularity of most videos is 5 or 6 fps.

To deeply analyze the influence of occlusion levels on model performance, OVIS provides the occlusion level annotation of every object in each frame. The occlusion levels are defined as follows: no occlusion, slight occlusion, and severe occlusion. As illustrated in Fig.~\ref{fig:various_occlusion}, no occlusion means the object is fully visible, slight occlusion means that more than 50\% of the object is visible, and severe occlusion means that more than 50\% of the object area is occluded. After the frame-level occlusion degree is annotated, we can quantify the occlusion degree of each instance through the whole video by gathering the occlusion level in all frames of the instance. Specifically, We first map the three occlusion levels mentioned before into numeric scores. The no occlusion, slight occlusion, and server occlusion are mapped into $0$, $0.25$, and $0.75$, respectively. Then, given an instance that appears in multiple frames, we use the averaged occlusion scores of top 50\% frames with highest scores to represent the occlusion degree of instances.

Each video is handled by one annotator to get the initial annotation, and the initial annotation is then passed to another annotator to check and correct if necessary. The final annotations will be examined by our research team and sent back for revision if deemed below the required quality.

While being designed for video instance segmentation, it should be noted that OVIS is also suitable for evaluating video object segmentation in either a semi-supervised or unsupervised fashion, and object tracking since the bounding-box annotation is also provided. The relevant experimental settings will be explored as part of our future work.

\subsection{Dataset Statistics}

\begin{table}
\centering
  \begin{tabular}{|l|r|r|p{1.1cm}<{\raggedleft}|}
  \hline
  Dataset & YTVIS 19 & YTVIS 21 & OVIS \\
  \hline
  \hline
  Masks & 131k & 232k & 296k \\ 
  Instances & 4,883 & 8,171 & 5,223 \\
  Categories & 40 & 40 & 25 \\   
  Videos & 2,883 & 3,859 & 901 \\
  Video duration$^\star$ & 4.61s & 5.03s & 12.77s \\ 
  Instance duration & 4.47s & 4.73s & 10.05s \\ 
  \hline
  mBOR$^\star$ & 0.07 & 0.06 & 0.22 \\
  Objects / frame$^\star$  & 1.57 & 1.95 & 4.72\\ 
  Instances / video$^\star$ & 1.69 & 2.10 & 5.80\\ 
  \hline
  \end{tabular}
\caption{Comparing OVIS with YouTube-VIS in terms of statistics. See Eq.~\eqref{eq:BOR} for the definition of mBOR. $\star$ means the value of YouTube-VIS is estimated from the training set.}
\label{tb:statistics}
\end{table}

\begin{figure*}
\centering
  \subfigure[]
  {
      \includegraphics[width=0.45\linewidth,height=0.27\linewidth]{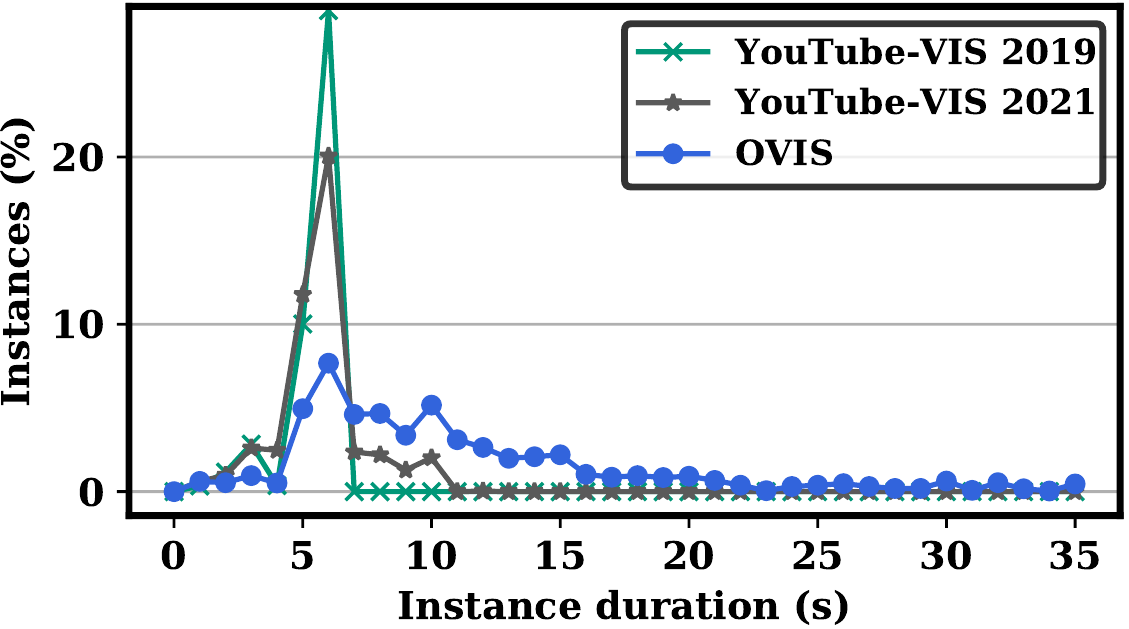}
      \label{fig:length_of_instacne_distribution}
  }
  \qquad
  \subfigure[]
  {
      \includegraphics[width=0.45\linewidth,height=0.27\linewidth]{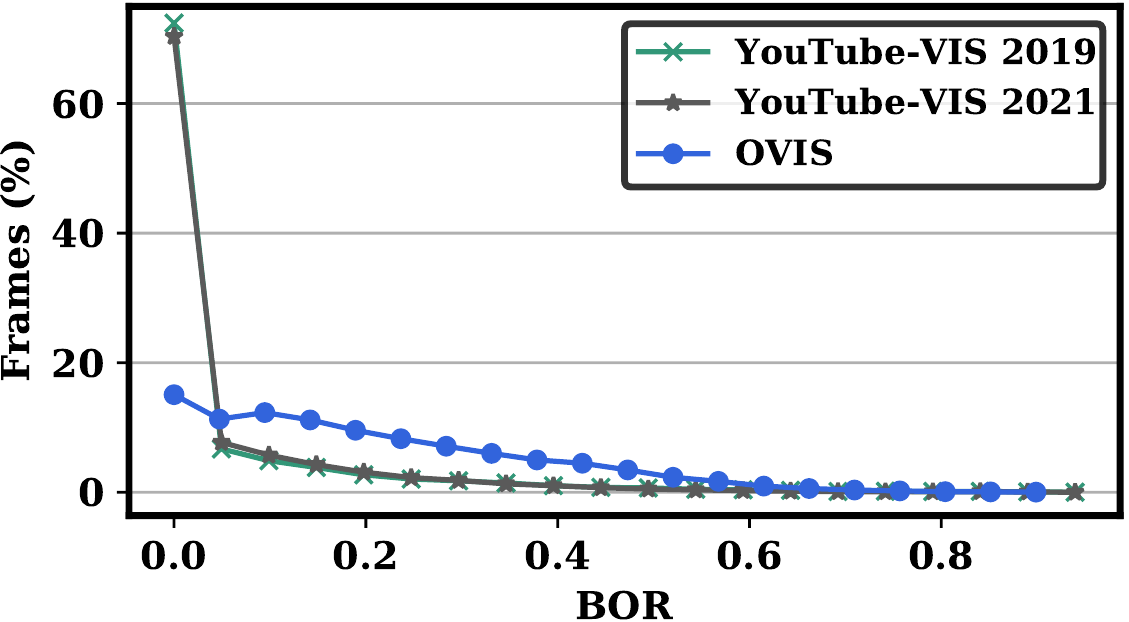}
      \label{fig:imi_distribution}
  }
  \subfigure[]
  {
      \includegraphics[width=0.45\linewidth,height=0.27\linewidth]{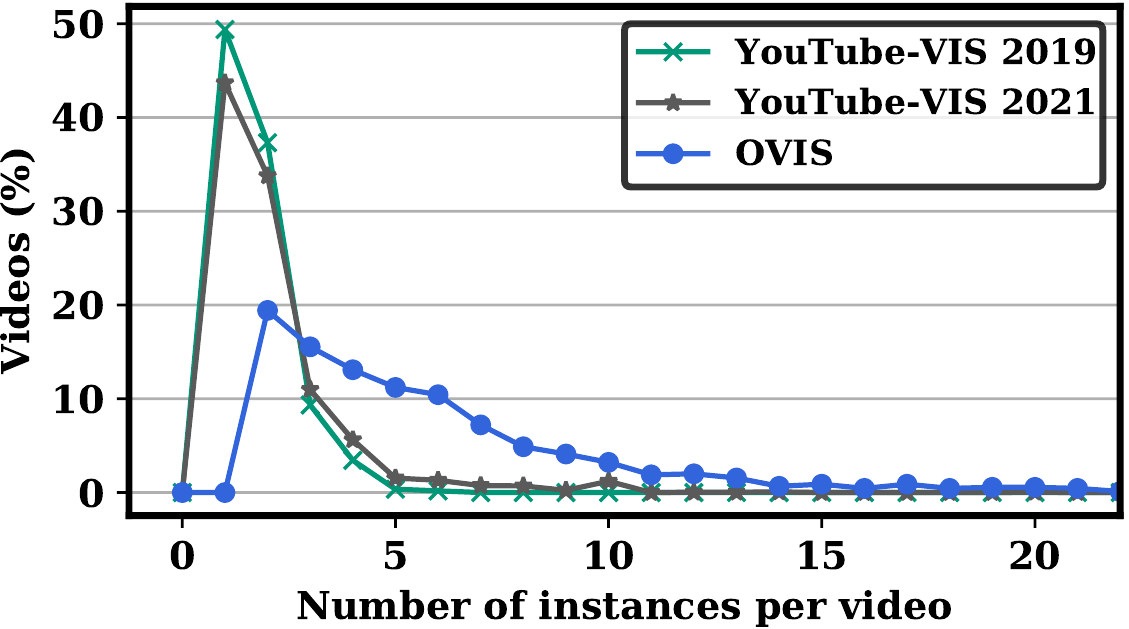}
      \label{fig:instances_per_video_distribution}
  }
  \qquad
  \subfigure[]
  {
      \includegraphics[width=0.45\linewidth,height=0.27\linewidth]{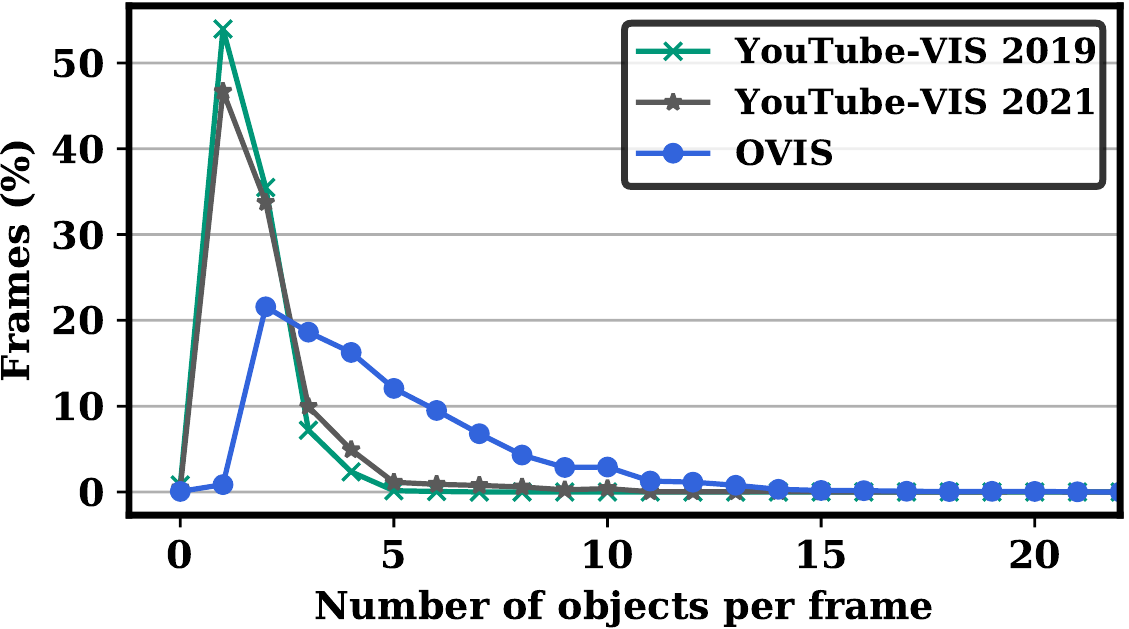}
      \label{fig:objects_per_frame_distribution}
  }
   \caption{Comparison of OVIS with YouTube-VIS, including the distribution of instance duration (a), BOR (b), the number of instances per video (c), and the number of objects per frame (d).}
   \label{fig:instance_density}
\end{figure*}

As YouTube-VIS~\cite{youtube_vis} is the only dataset that is specifically designed for video instance segmentation nowadays, we analyze the data statistics of OVIS with YouTube-VIS as a reference in Table~\ref{tb:statistics}. We compare OVIS with two versions of YouTube-VIS: YouTube-VIS 2019 and YouTube-VIS 2021. Note that some statistics, marked with $\star$, of YouTube-VIS are only calculated from the training set because only the annotation of the training set is publicly available. Nevertheless, considering the training set occupies 78\% of the whole dataset, those statistics could still reflect the properties of YouTube-VIS roughly.

In terms of basic and high-level statistics, OVIS contains 296k masks and 5,223 instances. The number of masks in OVIS is larger than YouTube-VIS 2019 and YouTube-VIS 2021 that have 131k and 232k masks, respectively. The number of instances in OVIS is larger than YouTube-VIS 2019 that has 4,883 instances, and less than YouTube-VIS 2021 that has 8,171 instances. Note that there are fewer categories in OVIS, so the mean instances count per category is larger than YouTube-VIS 2021. Nonetheless, \textit{OVIS has fewer videos than YouTube-VIS as our design philosophy favors long videos and instances so as to preserve enough motion and occlusion scenarios.}

As is shown, the average video duration and the average instance duration of OVIS are 12.77s and 10.05s respectively. Fig.~\ref{fig:length_of_instacne_distribution} presents the distribution of instance duration, which shows that all instances in YouTube-VIS last less than 10s. Long videos and instances increase the difficulty of tracking and the ability of long-term tracking is required.

\begin{figure}[t]
\centering
         \includegraphics[width=0.95\linewidth]{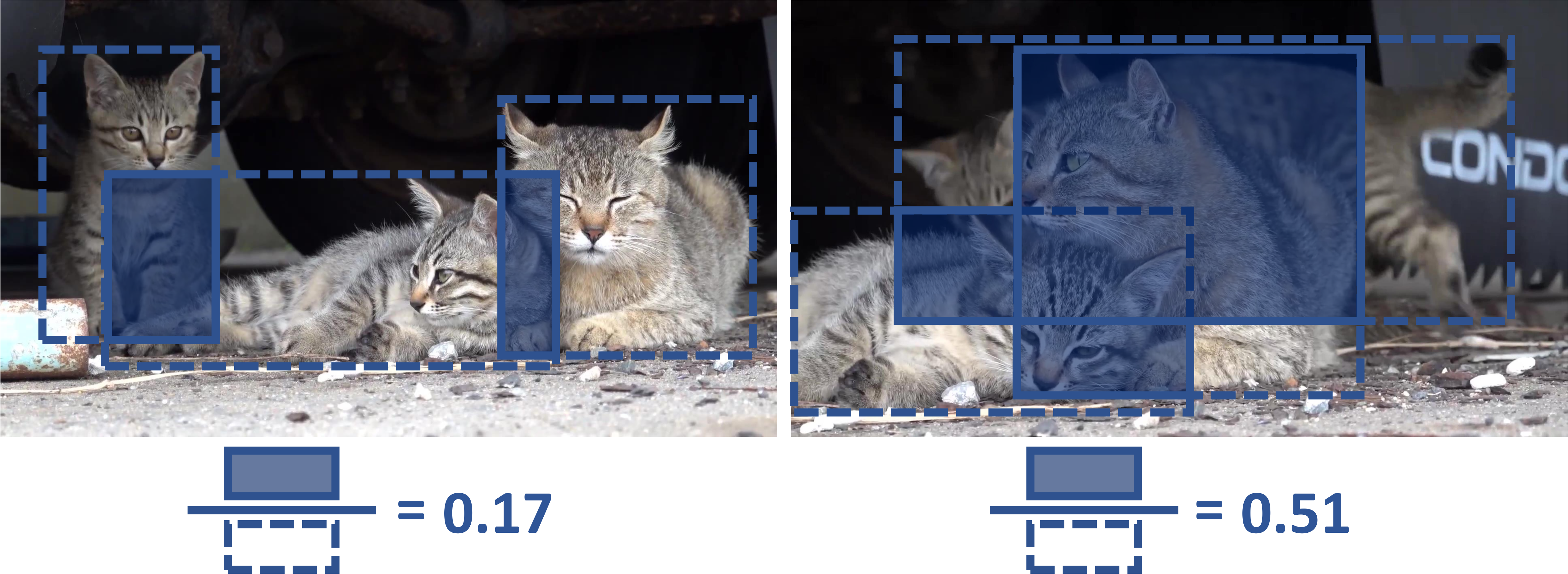}
         \caption{Visualization of occlusions with different BOR values. }
      \label{fig:bor}
\end{figure}

As for occlusion levels, the proportions of objects with no occlusion, slight occlusion, and severe occlusion in OVIS are 18.2\%, 55.5\%, and 26.3\% respectively. 80.2\% of instances are severely occluded in at least one frame, and only 2\% of the instances are not occluded in any frame. It supports the focus of our work, that is, to explore the ability of video instance segmentation models in handling occlusion scenes.

In order to compare the occlusion degree with other datasets, we define a metric named Bounding-box Occlusion Rate (BOR) to approximate the degree of occlusion. Given a video frame with $N$ objects denoted by bounding boxes $\{\textbf{B}_1,\textbf{B}_2,\dots,\textbf{B}_N\}$, we compute the BOR for this frame as
\begin{equation}
\label{eq:BOR}
\text{BOR}=\frac{|\bigcup_{1\leq i<j\leq N} \textbf{B}_i\bigcap\textbf{B}_j| }{|\bigcup_{1\leq i\leq N}\textbf{B}_i| },
\end{equation}
where the numerator means the area sum of the intersection between any two or more bounding boxes.
The denominator means the area of the union of all the bounding boxes. An illustration is given in Fig.~\ref{fig:bor}, which shows the larger the BOR value is, the heavier the occlusion is. 

Then we utilize mBOR, the average value of BORs of all the frames in a dataset (frames that do not contain any objects are ignored), to characterize the dataset in terms of the occlusion. As shown in Table~\ref{tb:statistics}, the mBOR of OVIS is 0.22, much higher than that of YouTube-VIS 2019 and YouTube-VIS 2021 (0.07 and 0.06, respectively). The BOR distribution is further compared in Fig.~\ref{fig:imi_distribution}. As can be seen, most frames in YouTube-VIS are located in the region where $\text{BOR}\leq0.1$. In comparison, the BOR of about half frames in OVIS is no less than 0.2. This supports that there are more severe occlusions in OVIS than YouTube-VIS.
However, it should be mentioned here that BOR can only roughly reflect the occlusion between objects. Therefore, \textit{mBOR could serve as an effective indicator for occlusion degrees, but only reflect the occlusion degree in a partial or rough way}. 

In addition to long videos\&instances and severe occlusions, OVIS features crowded scenes, which is a natural result caused by heavy occlusions. OVIS has 5.80 instances per video and 4.72 objects per frame, while those two values are 2.10 and 1.95 respectively in YouTube-VIS 2021. The comparison of the two distributions is further depicted in Fig.~\ref{fig:instances_per_video_distribution} and Fig.~\ref{fig:objects_per_frame_distribution}. 

\subsection{Evaluation Metrics}
\label{metric}

Following previous methods~\cite{youtube_vis}, we use
average precision (AP) at different intersection-over-union
(IoU) thresholds and average recall (AR) as the evaluation
metrics. The mean value of APs is also employed.

In addition, thanks to the occlusion level annotations in OVIS, we are able to analyze the performance under different occlusion levels. We divide all instances into three groups called slightly occluded, moderately occluded, and heavily occluded, in which the occlusion scores of instances are in the range of $[0, 0.25]$, $[0.25, 0.5]$, $[0.5, 0.75]$ respectively. The proportions of the three groups are 23\%, 44\%, and 49\% respectively. Then, we can get the AP of each group (denoted by $\text{AP}_{SO}$, $\text{AP}_{MO}$, and $\text{AP}_{HO}$ respectively) by ignoring the instances of other groups.

\section{Experiments} \label{section:experiment}
In this section, we comprehensively study the newly collected OVIS dataset by conducting experiments on 9 existing video instance segmentation algorithms and propose our new baseline method.
 
\subsection{Implementation Details}
\label{sec:implementation}

\paragraph{Datasets.}~On the newly collected OVIS dataset, the whole dataset is divided into 607 training videos, 140 validation videos, and 154 test videos. The split proportions of different categories are approximately the same, and there are at least 4 videos per category in the validation and test set. 
This split will be fixed as an official split. If not specified, the experiments are conducted on the validation set of OVIS.

\begin{table*}[!t]
\centering
\resizebox{\textwidth}{!}{
\begin{tabular}{|l|p{0.48cm}<{\centering}p{0.48cm}<{\centering}p{0.48cm}<{\centering}p{0.48cm}<{\centering}p{0.61cm}<{\centering}|p{0.61cm}<{\centering}p{0.61cm}<{\centering}p{0.61cm}<{\centering}|p{0.48cm}<{\centering}p{0.48cm}<{\centering}p{0.48cm}<{\centering}p{0.48cm}<{\centering}p{0.61cm}<{\centering}|p{0.61cm}<{\centering}p{0.61cm}<{\centering}p{0.61cm}<{\centering}|}
\hline
\multirow{2}{*}{Methods} & \multicolumn{8}{c|}{OVIS validation set}  & \multicolumn{8}{c|}{OVIS test set} \\
\cline{2-9} \cline{10-17}
& AP  & $\text{AP}_{50}$  & $\text{AP}_{75}$  & $\text{AR}_{1}$  & $\text{AR}_{10}$  & $\text{AP}_{SO}$  & $\text{AP}_{MO}$  & $\text{AP}_{HO}$  & AP  & $\text{AP}_{50}$ & $\text{AP}_{75}$  & $\text{AR}_{1}$  & $\text{AR}_{10}$  & $\text{AP}_{SO}$  & $\text{AP}_{MO}$  & $\text{AP}_{HO}$ \\
\hline
\hline
FEELVOS~\cite{feelvos} & 9.6 & 22.0 & 7.3 & 7.4 & 14.8 & 17.3 & 11.5 & 1.7 & 10.8 & 23.4 &8.7 &9.0 & 16.2 & 18.9 & 12.2 & 2.0 \\ 
IoUTracker+~\cite{youtube_vis} & 7.0 & 16.9 & 5.3 & 5.7 & 14.3 & 11.5 & 7.9 & 1.8 & 8.0 & 18.4 & 7.5 & 5.9 & 15.7 & 12.8 & 9.1 & 2.1 \\
\small{MaskTrack R-CNN~\cite{youtube_vis}} & 10.8 & 25.3 & 8.5 & 7.9 & 14.9 & 23.0 & 12.8 & 2.7 & 11.8 & 25.4 & 10.4 & 7.9 & 16.0 & 22.7 & 15.0 & 3.5 \\
SipMask~\cite{sipmask} & 10.2 & 24.7 & 7.8 & 7.9 & 15.8 & 19.9 & 10.5 & 2.2 & 11.7 & 23.7  &10.5 & 8.1 & 16.6 & 21.9 & 13.9 & 3.2 \\
STEm-Seg~\cite{stem_seg}  & 13.8 & 32.1 & 11.9 & 9.1 & 20.0 & 22.2 & 16.1 & 3.9 & 14.4 &30.0 & 13.0 & 10.1 & 20.6 & 22.5 & 16.8 & 4.2 \\
TraDeS~\cite{trades}  & 11.4 & 26.5 & 9.4 & 7.0 & 13.8 & 23.0 & 12.8 & 3.0 & 12.0 & 26.4 & 10.8 & 7.8 & 14.6 & 21.6 & 14.1 & 3.6 \\
QueryVIS~\cite{queryinst}  & 14.7 & \textbf{34.7} & 11.6 & 9.0 & 21.2 & 27.3 & 17.2 & 4.1 & 16.0 & \textbf{33.7} & 14.7 & 9.6 & 21.7 & 26.3 & 17.7 & 4.5 \\
STMask~\cite{stmask}  & \textbf{15.4} & 33.8 & \textbf{12.5} & 8.9 & \textbf{21.3} & 24.0 & \textbf{18.7} & \textbf{5.1} & 15.6 & 32.5 & 13.8 & 9.1 & \textbf{21.8} & 25.4 & 17.1 & \textbf{6.3} \\
CrossVIS*~\cite{crossvis}  & 14.9 & 32.7 & 12.1 & \textbf{10.3} & 19.8 & \textbf{28.4} & 16.9 & 4.1 & \textbf{16.3} & 31.5 & \textbf{15.4} & \textbf{10.6} & 21.1 & \textbf{27.3} & \textbf{18.5} & 5.6 \\
\hline
\end{tabular}}
\caption{Overall results of state-of-the-art methods on the OVIS dataset. $\text{AP}_{SO}$, $\text{AP}_{MO}$, and $\text{AP}_{HO}$ respectively denote the AP of ``slightly occluded", ``moderately occluded", and ``heavily occluded". * means the baseline model is additionally pre-trained with the YouTube-VIS dataset~\cite{youtube_vis}.}
\label{main_results}
\end{table*}

\begin{figure}[!t]
\centering
\includegraphics[width=0.95\linewidth]{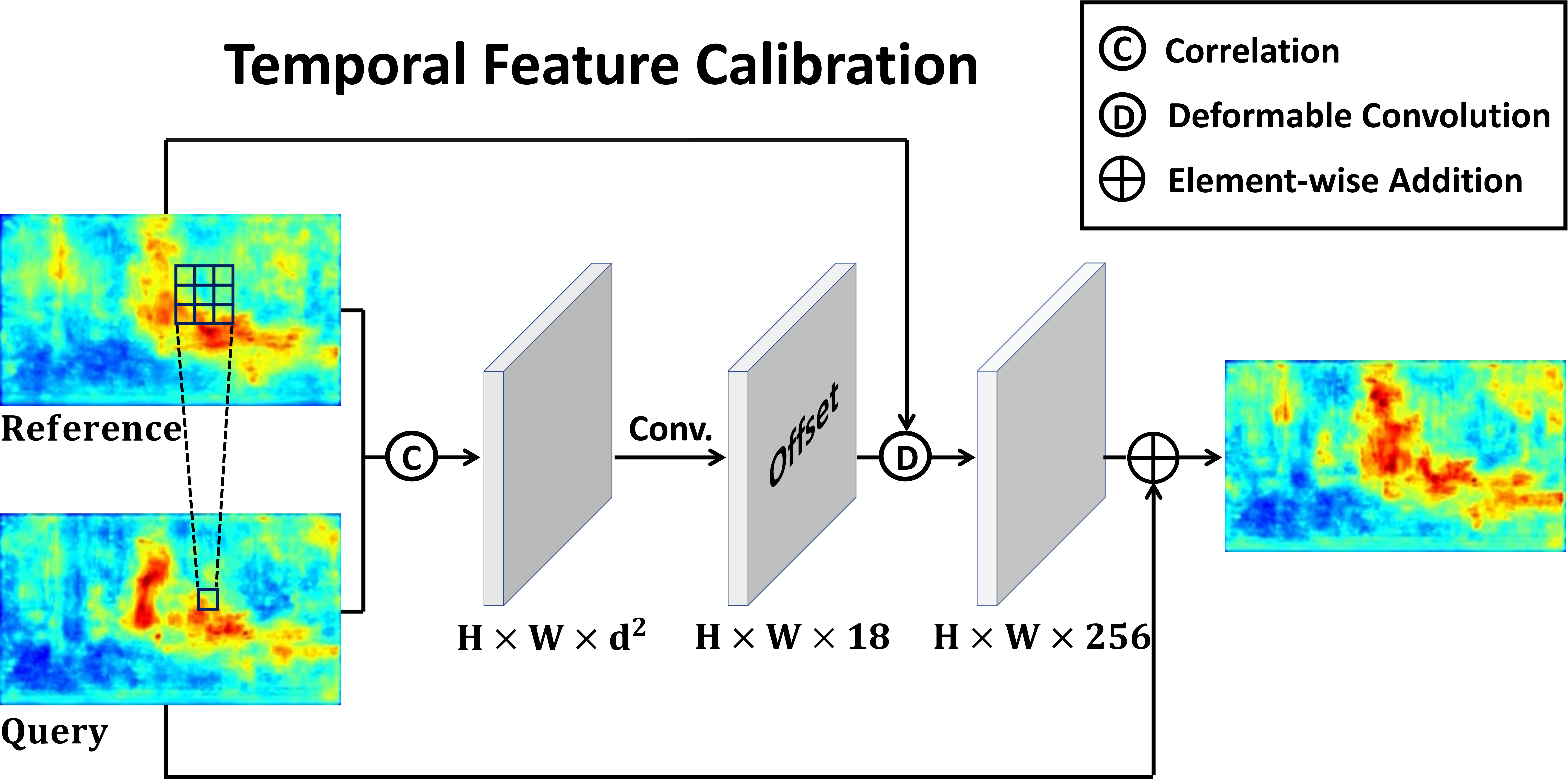}
\caption{The pipeline of temporal feature calibration, which can be inserted into different video instance segmentation models by changing the following prediction head.}
\label{fig:overall_architecture}
\end{figure}

\paragraph{A Temporal Feature Calibration Plug-in.}~One of the keys to tackling occlusion is to complement the missing object cues. In a video that has a temporal dimension, a mild assumption is that usually, the missing object cues in the current frame may have appeared in adjacent frames. Hence, it is natural to leverage adjacent frames to alleviate occlusions. However, caused by motions, the features of different frames are not aligned in the spatial dimension. Things get much worse because of the existence of severe occlusions. To solve this issue, following \cite{spatiotemporal_sampling,flownet}, we present an easy plug-in called temporal feature calibration as illustrated in Fig.~\ref{fig:overall_architecture}.

Denote by $\textbf{F}_\textbf{q}\in\mathbb{R}^{H\times W\times C}$ and $\textbf{F}_\textbf{r}\in\mathbb{R}^{H\times W\times C}$ the feature tensor of the query frame (called target or current frame in some literature) and a reference frame, respectively. The feature calibration first computes the spatial correlation~\cite{flownet} between $\textbf{F}_\textbf{q}$ and $\textbf{F}_\textbf{r}$. Given a location $\textbf{x}_\textbf{q}$ in $\textbf{F}_\textbf{q}$ and $\textbf{x}_\textbf{r}$ in $\textbf{F}_\textbf{r}$, we compute 
\begin{equation}
\textbf{c}\textbf{(}\textbf{x}_\textbf{q},\textbf{x}_\textbf{r}\textbf{)} = \sum_{o\in[-k,k]\times[-k,k]}\textbf{F}_\textbf{q}(\textbf{x}_\textbf{q}+o)\textbf{F}_\textbf{r}(\textbf{x}_\textbf{r}+o)^\mathrm{T}\text{.}
\end{equation}
The above operation will transverse the $d\times d$ area centered on $\textbf{x}_\textbf{q}$, then outputs a $d^2$-dimensional vector.

After enumerating all the positions in $\textbf{F}_\textbf{q}$, we obtain $\textbf{C}\in\mathbb{R}^{H\times W\times d^2}$ and forward it into multiple stacked convolution layers to get the spatial calibration offset $\textbf{D}\in\mathbb{R}^{H\times W\times 18}$. We then obtain a calibrated version of $\textbf{F}_\textbf{r}$ by applying deformable convolutions with $\textbf{D}$ as the spatial calibration offset, which is denoted as $\overline{\textbf{F}}_{\textbf{r}}$. At last, we fuse the calibrated reference feature $\overline{\textbf{F}}_{\textbf{r}}$ with the query feature $\textbf{F}_\textbf{q}$ by element-wise addition for the localization, classification, and segmentation of the current frame afterward.

During training, for each query frame $\textbf{F}_\textbf{q}$, we randomly sample a reference frame $\textbf{F}_\textbf{r}$ from the same video. As compared with the short videos in YouTube-VIS (the longest video in YouTube-VIS contains only 36 frames), the first frame and the last frame of a long video in OVIS (the longest video in OVIS contains 500 frames) may be totally different. In order to ensure that the reference frame has a strong spatial correspondence with the query frame, the sampling is only done locally within $\epsilon_\text{train}=5$ frames. Since the temporal feature calibration is differentiable, it can be trained end-to-end by the original detection and segmentation loss. When inference, all frames adjacent to the query frame within the range $\epsilon_\text{test}=5$ are taken as reference frames. We linearly fuse the classification confidences, regression bounding box coordinates, and segmentation masks obtained from each reference frame and output the final results for the query frame.

In the experiments, we denote the new methods as CMaskTrack R-CNN and CSipMask, when Calibrating MaskTrack R-CNN~\cite{youtube_vis} models and Calibrating SipMask~\cite{sipmask} models, respectively.

\paragraph{Experimental Setup.}~For all our experiments, we adopt ResNet-50-FPN~\cite{resnet} as the backbone. The models are initialized by Mask R-CNN which is pre-trained on MS-COCO~\cite{coco}. All frames are resized to $640 \times 360 $ during both training and inference for fair comparisons with previous works~\cite{youtube_vis,sipmask,stem_seg}.
For our new baselines (CMaskTrack R-CNN and CSipMask), we use three convolution layers of kernel size $3\times3$ in the module for temporal feature calibration. The training epoch is set to 12, and the initial learning rate is set to 0.005 and decays with a factor of 10 at epoch 8 and 11.

\subsection{Main Results}

On the OVIS dataset, we first produce the performance of several state-of-the-art algorithms whose code is publicly available, including mask propagation methods (\eg, FEELVOS~\cite{feelvos}), track-by-detect methods (\eg,~IoUTracker+~\cite{youtube_vis}), and recently proposed end-to-end methods (\eg,~MaskTrack R-CNN~\cite{youtube_vis}, SipMask~\cite{sipmask}, STEm-Seg~\cite{stem_seg}, STMask~\cite{stmask}, TraDeS~\cite{trades}, CrossVIS~\cite{crossvis}, and QueryVIS~\cite{queryinst}). \myTextColor{The standard deviation of the reported results below is about 0.5.}

As presented in Table~\ref{main_results}, although most of these methods can obtain more than 30 AP on the YouTube-VIS dataset, all of them encounter a great performance degradation of at least 50\% on OVIS compared with that on YouTube-VIS. Especially in the heavily occluded instance group, all methods suffer from a significant performance drop of more than 80\%. For example, SipMask~\cite{sipmask}, which achieves an AP of 32.5 on YouTube-VIS, only obtains an AP of 2.2 in the heavily occluded group of OVIS validation set. It firmly suggests that severe occlusion will greatly improve the difficulty of video instance segmentation, and further attention should be paid to video instance segmentation in the real world where occlusions extensively happen.
Benefiting from the feature calibration and temporal fusion, STMask~\cite{stmask} obtains an $\text{AP}_{HO}$ of 5.1 on the validation set and 6.3 on the test set, surpassing all other methods in the heavily occluded group.

\begin{table*}[!t]
\centering
\begin{tabular}{|l|p{0.55cm}<{\centering}p{0.48cm}<{\centering}p{0.56cm}<{\centering}p{0.47cm}<{\centering}p{0.61cm}<{\centering}|p{0.61cm}<{\centering}p{0.61cm}<{\centering}p{0.61cm}<{\centering}|p{0.48cm}<{\centering}p{0.48cm}<{\centering}p{0.48cm}<{\centering}p{0.47cm}<{\centering}p{0.61cm}<{\centering}|}
\hline
\multirow{2}{*}{Methods} & \multicolumn{8}{c|}{OVIS validation set} & \multicolumn{5}{c|}{YouTube-VIS 2019 validation set}  \\
\cline{2-9} \cline{10-14}
& AP  & $\text{AP}_{50}$  & $\text{AP}_{75}$  & $\text{AR}_{1}$  & $\text{AR}_{10}$ & $\text{AP}_{SO}$  & $\text{AP}_{MO}$  & $\text{AP}_{HO}$  & AP  & $\text{AP}_{50}$ & $\text{AP}_{75}$  & $\text{AR}_{1}$  & $\text{AR}_{10}$ \\
\hline
\hline
SipMask~\cite{sipmask} & 10.2 & 24.7 & 7.8 & 7.9 & 15.8 & 19.9 & 10.5 & 2.2 & 32.5 & 53.0 & 33.3 & 33.5 & 38.9\\
CSipMask  & 14.3 & 29.9 & 12.5 & \textbf{9.6} & 19.3 & 27.1 & 16.6 & 3.2 &\textbf{35.1} & \textbf{55.6} &\textbf{38.1} &\textbf{35.8}  &\textbf{41.7} \\
\hline
\small{MaskTrack R-CNN~\cite{youtube_vis}} & 10.8 & 25.3 & 8.5 & 7.9 & 14.9 & 23.0 & 12.8 & 2.7 & 30.3 & 51.1 & 32.6 & 31.0 & 35.5 \\
CMaskTrack R-CNN & \textbf{15.4} & \textbf{33.9} & \textbf{13.1} & 9.3 & \textbf{20.0} &\textbf{28.6} & \textbf{18.7} & \textbf{4.1} & 32.1 & 52.8 & 34.9 & 33.2 & 37.9\\
\hline
\end{tabular}
\caption{Quantitative comparison between the new methods and their corresponding baselines on the OVIS dataset and the YouTube-VIS dataset.}
\label{tb:tfc_ablation}
\end{table*}

\begin{figure*}[t]
\centering
  \includegraphics[width=0.95\linewidth]{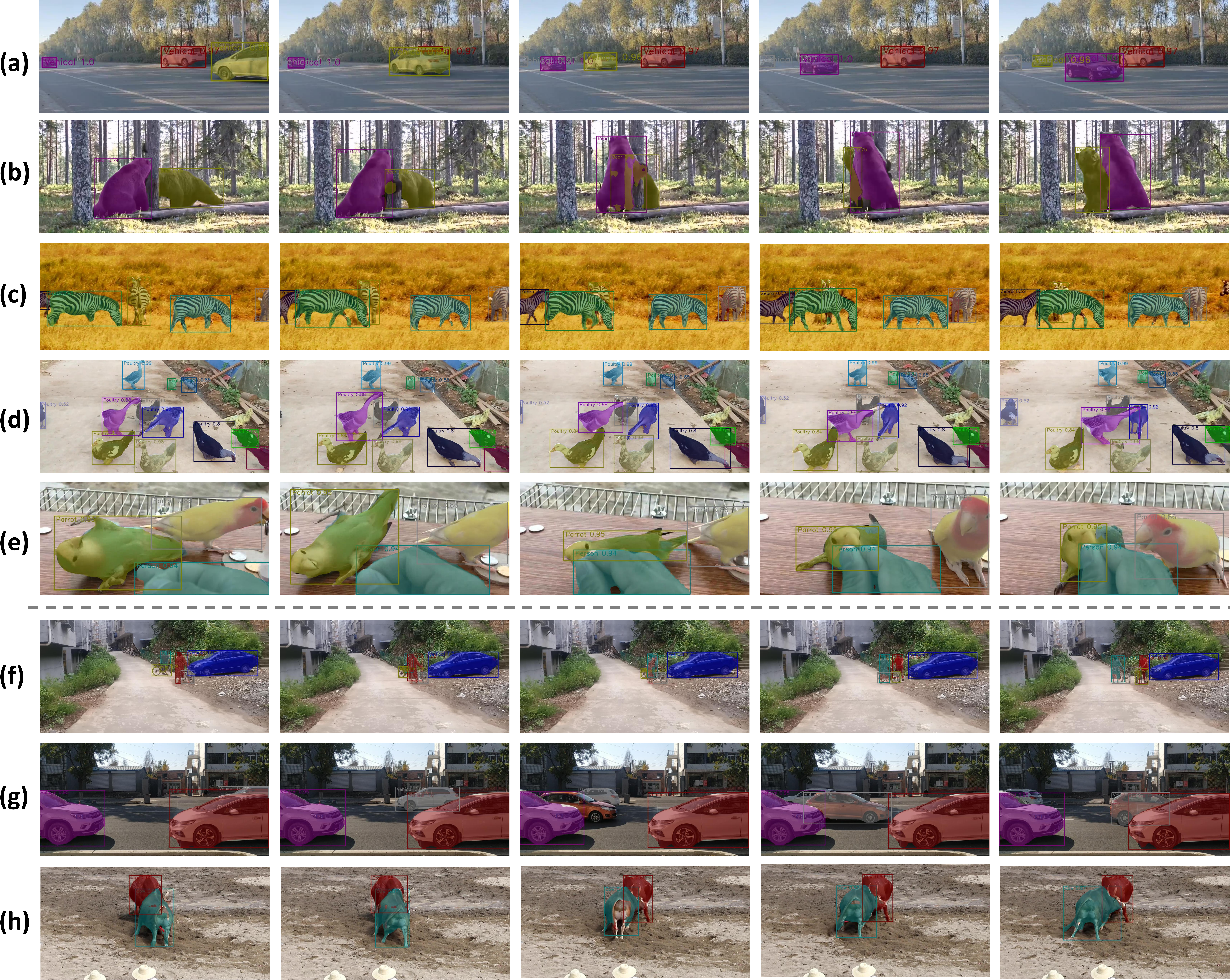}
  \caption{Evaluation examples on OVIS. Each row presents the results of 5 frames in a video sequence. (a)-(e) are successful cases and (f) and (h) are failure cases.}
\label{fig:frames_Visilation}
\end{figure*}

\myTextColor{It is worth noting that, as the only bottom-up video instance segmentation method, STEm-Seg achieves similar $\text{AP}_{SO}$ with MaskTrack R-CNN and TraDeS, but much higher $\text{AP}_{HO}$ (3.9 \vs 2.7 \vs 3.0). It demonstrates that the bottom-up paradigm like STEm-Seg may perform better than the general top-down paradigm on occlusion handling. Our interpretation is that the bottom-up architecture avoids the detection process which is difficult in occluded scenes.}

In addition, as shown in Table~\ref{tb:tfc_ablation}, by leveraging the feature calibration module, the performance on OVIS is significantly improved. CMaskTrack R-CNN leads to an AP improvement of 4.6 over MaskTrack R-CNN  (10.8~\textit{vs.}~15.4), and CSipMask leads to an AP improvement of 4.1 over SipMask (10.2~\textit{vs.}~14.3). Besides, the experiments also show that TFC can boost the performance of all occlusion levels, and the improvement of heavy occlusion and moderate occlusion is more significant (see Fig.~\ref{fig:epsilon_test_occlusion_level} for more details). We also evaluate the proposed CMaskTrack R-CNN and CSipMask on the YouTube-VIS dataset. As shown in Table~\ref{tb:tfc_ablation}, CMaskTrack R-CNN and CSipMask surpass the corresponding baselines by 1.8 and 2.6 in terms of AP, respectively, which demonstrates the flexibility and the generalization power of the proposed feature calibration module.

To present the qualitative evaluation results of methods on OVIS, some evaluation examples of CMaskTrack R-CNN are given in Fig.~\ref{fig:frames_Visilation}, including 5 successful cases (a)-(e) and 3 failure cases (f) and (h). In (a), the car in the yellow mask first blocks the car in the red mask entirely in the 2nd frame, then is entirely blocked by the car in the purple mask in the 4th frame. It is surprising that even in this extreme case, all the cars are well tracked. In (b), CMaskTrack R-CNN successfully tracks the bear in the yellow mask, which is partially occluded by another object,~\ie,~the bear in the purple mask, and the background,~\ie,~the tree. In (d), we present a crowded scene where almost all the ducks are correctly detected and tracked. In (f), two persons and two bicycles heavily overlap with each other. CMaskTrack R-CNN fails to track the person and segment the bicycle. In (g), when two cars are intersecting, severe occlusion leads to failure of detection and tracking. In (h), although humans could sense that there are two persons with hats at the bottom, CMaskTrack R-CNN cannot detect and track them because the appeared visual cues are inadequate.

\subsection{Discussions}

\begin{table}
\centering
\resizebox{\linewidth}{!}{
\begin{tabular}{|l|l|cc|}
\hline
\multicolumn{2}{|c|}{Dataset} & OVIS & \small{YouTube-VIS~\cite{youtube_vis}}\\
\hline
\hline
\multirow{2}{*}{Image Oracle} 
&AP & 58.4 \scriptsize{($\uparrow$441\%)} & 78.7 \scriptsize{($\uparrow$160\%)} \\ 
&$\text{AR}_{10}$ & 66.1 \scriptsize{($\uparrow$460\%)} & 83.7 \scriptsize{($\uparrow$136\%)} \\
\hline
\multirow{2}{*}{\small{Identity Oracle}} 
&AP & 23.9 \scriptsize{($\uparrow$121\%)} & 31.5 \scriptsize{($\uparrow$\hspace{2.9mm}4\%)} \\ 
&$\text{AR}_{10}$ & 28.2 \scriptsize{($\uparrow$139\%)} & 34.6 \scriptsize{($\uparrow$\hspace{2.3mm}-2\%)} \\
\hline
\end{tabular}}
\caption{Oracle results on OVIS and YouTube-VIS. The number in the brackets means the performance improvement ratio over the corresponding baseline.}
\label{tb:oracle}
\end{table}

\paragraph{Oracle Results.}~We conduct the image oracle and identity oracle experiments to explore the impact of image-level prediction and cross-frame association on the performance of the OVIS dataset. In order to compare with the YouTube-VIS dataset~\cite{youtube_vis}, we use MaskTrack-RCNN for experiments. Following~\cite{youtube_vis}, in the image oracle experiments, we use ground-truth bounding boxes, masks, and category labels to replace the predictions by MaskTrack R-CNN, and then track those ground-truth bounding boxes by the tracking branch. In the identity oracle experiment, we first assign each per-frame prediction to the closest ground-truth bounding box, and then aggregate the bounding boxes with the same identity through the video.

The results are shown in Table~\ref{tb:oracle}. On the OVIS dataset, the image oracle experiments and identity oracle experiments obtain 58.4 and 25.5 AP, respectively. \myTextColor{This demonstrates that the image level prediction is more critical for the performance of occluded video instance segmentation, which is mainly associated to object segmentation and classification in frames. It can be expected that more advanced image-based techniques could be explored further so as to approach this upper limit.} Interestingly, both oracle experiments achieve lower performance on the OVIS dataset than that on the YouTube-VIS dataset, which shows that whether for image-level prediction or cross-frame association, the OVIS dataset is more challenging than the YouTube-VIS dataset.
Moreover, in identity experiments, the AP on YouTube-VIS achieves almost no gain (only 4\% improvement over the result of MaskTrack R-CNN baseline), while the AP on OVIS is greatly improved (121\% improvement over MaskTrack R-CNN), which demonstrates that the tracking task on OVIS is much more difficult than that on YouTube-VIS.

\paragraph{Effect of Leveraging Image Datasets.}~Caused by the high cost of exhaustively annotating high-quality video segmentation masks, video inadequacy is a common problem among existing video segmentation datasets. The lack of diversity in video scenes may affect the generalization capability of models trained on those datasets. To this end, we further train several models with both the video data in OVIS and additional augmented image sequences/pairs synthesized from other large-scale image instance segmentation datasets. In our experiments, the proportions of video data and augmented image data are 65\% and 35\% respectively. These pseudo image sequences are generated from the COCO~\cite{coco} dataset by on-the-fly random perspective and affine transformation. The evaluation results are shown in Table~\ref{tb:withimage}. We can see that by leveraging the augmented image sequences, all these three baseline methods can achieve remarkable AP improvements, which can serve as a reference for future research.

\begin{table*}
\centering
\begin{tabular}{|l|c|p{0.48cm}<{\centering}p{0.48cm}<{\centering}p{0.57cm}<{\centering}p{0.48cm}<{\centering}p{0.6cm}<{\centering}|p{0.6cm}<{\centering}p{0.6cm}<{\centering}p{0.6cm}<{\centering}|}
\hline
Methods & w/ image data & AP  & $\text{AP}_{50}$  & $\text{AP}_{75}$  & $\text{AR}_{1}$  & $\text{AR}_{10}$ & $\text{AP}_{SO}$  & $\text{AP}_{MO}$  & $\text{AP}_{HO}$ \\
\hline
\hline
\multirow{2}{*}{MaskTrack R-CNN~\cite{youtube_vis}}        & & 10.8 & 25.3 & 8.5 & \textbf{7.9} & 14.9 & 23.0 & 12.8 & 2.7\\
& \checkmark & \textbf{12.0} & \textbf{28.5} & \textbf{8.9} & 7.8 & \textbf{16.3} & \textbf{24.7} & \textbf{14.2} & \textbf{3.2}\\
\hline
\multirow{2}{*}{SipMask~\cite{sipmask}}        & & 10.2 & 24.7 & 7.8 & 7.9 & 15.8 & 19.9 & 10.5 & 2.2\\
& \checkmark & \textbf{11.8} & \textbf{27.5} & \textbf{9.0} & \textbf{8.3} & \textbf{17,7} & \textbf{23.2} & \textbf{13.4} & \textbf{2.3}\\
\hline
\multirow{2}{*}{STEm-Seg~\cite{stem_seg}}      & & 13.8 & 32.1 & 11.9 & 9.1 & 20.0 & 22.2 & 16.1 & 3.9\\
& \checkmark & \textbf{15.2} & \textbf{34.7} & \textbf{12.5} & \textbf{10.7} & \textbf{23.7} & \textbf{25.5} & \textbf{17.4} & \textbf{4.1}\\
\hline
\end{tabular}
\caption{The results of training with and without augmented image sequences. ``w/ image data" means training with both video data and the synthesized clips.}
\label{tb:withimage}
\end{table*}

\begin{table*}
\centering
\begin{tabular}{|l|p{0.48cm}<{\centering}p{0.48cm}<{\centering}p{0.57cm}<{\centering}p{0.48cm}<{\centering}p{0.6cm}<{\centering}|p{0.6cm}<{\centering}p{0.6cm}<{\centering}p{0.6cm}<{\centering}|}
\hline
Methods & AP  & $\text{AP}_{50}$  & $\text{AP}_{75}$  & $\text{AR}_{1}$  & $\text{AR}_{10}$ & $\text{AP}_{SO}$  & $\text{AP}_{MO}$  & $\text{AP}_{HO}$ \\
\hline
\hline
MaskTrack R-CNN & 10.8 & 25.3 & 8.5 & 7.9 & 14.9 & \textbf{23.0} & 12.8 & 2.7 \\
$+$ Adaptive NMS~\cite{adaptivenms} Oracle & \textbf{11.2} & \textbf{26.5} & \textbf{8.6} & \textbf{8.3} & \textbf{15.6} & 22.8 & \textbf{13.0} & \textbf{2.9} \\
\hline
\end{tabular}
\caption{\myTextColor{Adaptive NMS oracle results of MaskTrack R-CNN on OVIS.}}
\label{tb:nms_oracle}
\end{table*}

\paragraph{\myTextColor{Analysis of NMS Threshold.}}~\myTextColor{
Non-Maximum Suppression (NMS) is a necessary post-processing for most detection methods.}
    
\myTextColor{
To test the impact of NMS threshold on occlusion handling, inspired by \cite{adaptivenms}, we design the adaptive NMS oracle experiment. Specifically, for each ground-truth bounding box, we calculate the maximum IoU $d$ between it and all other ground-truth boxes. Then, the NMS threshold of all the predicted boxes that correspond to this ground-truth box will be assigned as $\text{max}(d, 0.5)$. In this way, a larger NMS threshold will be applied to the predictions in dense scenes, which can prevent NMS from removing the true positives that are close to other ground-truth boxes.}

\myTextColor{
As presented in Table~\ref{tb:nms_oracle}, based on MaskTrack R-CNN, the adaptive NMS oracle experiment improves $\text{AP}_{MO}$ and $\text{AP}_{HO}$ by 0.2, which proves that using a higher NMS threshold adaptively improves the performance in occluded scenes. However, even though we exactly know the real density~\cite{adaptivenms} of boxes in the adaptive NMS oracle experiment, $\text{AP}_{SO}$ decreases from 23.0 to 22.8. The overall AP only improves from 10.8 to 11.2, which shows that the NMS threshold adjusting is not a bottleneck on OVIS.}

\myTextColor{
One interpretation is that adjusting the NMS threshold is more important for tasks that require detecting the amodal bounding boxes (additionally containing the occluded invisible parts), such as the full-body bounding boxes in crowded pedestrian detection datasets~\cite{citypersons,crowdhuman}. For two occluded objects, the IoU of amodal bounding boxes will be much higher than the IoU of the bounding boxes of only the visible parts (like the boxes in OVIS). In addition, some learnable NMS methods~\cite{learnablenms,adaptivenms} have also been proposed, and many new methods~\cite{detr,queryinst} based on set prediction even do not need NMS post-processing. These new methods require further exploration in OVIS.}

\begin{table*}[!t]
\centering
\begin{tabular}{|l|l|ccc|c|}
\hline
Error type & Methods & No occlusion (\%) & Slight occlusion (\%) & Severe occlusion (\%) & All (\%)  \\
\hline
\hline
\multirow{3}{*}{Cls. error rate} & MaskTrack R-CNN & 41.3\hspace{7.3mm} & 41.0\hspace{6.2mm} & 50.1\hspace{6.2mm} & 42.5\hspace{6.2mm} \\
& MaskTrack R-CNN + LSS + DCN & 30.2 \scriptsize{(-11.1)} & 32.9 \scriptsize{(-8.1)} & 45.8 \scriptsize{(-4.3)} & 34.5 \scriptsize{(-8.0)} \\
& CMaskTrack R-CNN (ours) & \textbf{27.5} \scriptsize{(-\hspace{1.2mm}2.7)} & \textbf{29.0} \scriptsize{(-3.9)} & \textbf{39.2} \scriptsize{(-6.6)} & \textbf{30.2} \scriptsize{(-4.3)} \\
\hline
\multirow{3}{*}{Seg. error rate} & MaskTrack R-CNN & \textbf{12.1}\hspace{7.3mm} & 25.6\hspace{6.2mm} & 34.1\hspace{6.2mm} & 28.3\hspace{6.2mm} \\
& MaskTrack R-CNN + LSS + DCN & 13.2 \scriptsize{(+\hspace{0.6mm}0.9)} & 25.1 \scriptsize{(-0.5)} & 33.2 \scriptsize{(-0.9)} & 28.0 \scriptsize{(-0.3)} \\
& CMaskTrack R-CNN (ours) & 13.0 \scriptsize{(-\hspace{1.2mm}0.2)} & \textbf{22.2} \scriptsize{(-2.9)} & \textbf{29.5} \scriptsize{(-3.7)} & \textbf{25.3} \scriptsize{(-2.7)} \\
\hline
\multirow{3}{*}{ID switch rate} & MaskTrack R-CNN & 18.6\hspace{7.7mm} & 22.5\hspace{6.2mm} & 32.6\hspace{6.2mm} & 22.9\hspace{6.2mm} \\
& MaskTrack R-CNN + LSS + DCN & 12.5 \scriptsize{(-\hspace{1.2mm}6.1)} & 16.1 \scriptsize{(-6.4)} & 26.1 \scriptsize{(-6.5)} & 16.8 \scriptsize{(-6.1)} \\
& CMaskTrack R-CNN (ours) & \textbf{11.2} \scriptsize{(-\hspace{1.2mm}1.3)} & \textbf{14.0} \scriptsize{(-2.1)} & \textbf{21.6} \scriptsize{(-4.5)} & \textbf{14.4} \scriptsize{(-2.4)} \\
\hline
\end{tabular}
\caption{Error analysis under different occlusion levels. LSS denotes the local sampling strategy and DCN means applying a deformable convolutional layer on the query frame itself. \myTextColor{For a certain row, the number in the brackets means the decrease of error rates over the row above.}}
\label{error_analysis}
\end{table*}

\paragraph{Error Analysis.}To explore the detailed influence of occlusion levels on video instance segmentation, in this subsection, we analyze the frame-level error rates of classification, segmentation, and tracking under different occlusion levels. A segmentation error refers to that the IoU between the predicted mask of an object and its ground-truth less than 0.5 and the tracking error is reflected by ID switch rate.

Formally, we denote the predicted masks and labels in all frames as 
$M=\{m_1,m_2,...,m_n\}$ and $Y=\{y_1,y_2,...,y_n\}$, respectively, where $n$ is the number of predictions. The corresponding matched ground-truth masks and labels as 
$M^*=\{m^*_1,m^*_2,...,m^*_n\}$ and $Y^*=\{y^*_1,y^*_2,...,y^*_n\}$, respectively.

Regarding classification error rates, we consider the predicted object whose IoU with its matched ground-truth is greater than 0.5, then count the proportion of classification errors among them, as

\begin{equation}
\label{eq:error_cls}
\text{E}_{cls}=\frac{|\{m_j |\text{IoU}(m_j,m^*_j)>0.5 \wedge  y_j\not= y^*_j \}|} {|\{m_i | \text{IoU}(m_i,m^*_i)>0.5\}|}.
\end{equation}

For segmentation error rates, following~\cite{tide}, we consider masks whose IoU with its matched ground-truth is greater than 0.1. A mask $m_i$ will be counted as a segmentation error if its IoU with the corresponding ground-truth $m^*_i$ is less than 0.5. Then the segmentation error rate is calculated as
\begin{equation}
\label{eq:error_seg}
\text{E}_{seg}=\frac{|\{m_j |0.1<\text{IoU}(m_j,m^*_j)<0.5\}|} {|\{m_i | \text{IoU}(m_i,m^*_i)>0.1\}|}.
\end{equation}

The ID switch rate refers to the ratio of ID switches in the tracking sequence of all instances. Following~\cite{mots}, the predicted ID of a ground-truth instance in a frame is defined as the tracking ID of the closest predicted mask. If the ID of a ground-truth instance is not equal to that of its latest tracked predecessor, it will be considered as an ID switch.

\myTextColor{Based on the error rates defined above, we further evaluate MaskTrack and CMaskTrack. In addition, we define a baseline named ``MaskTrack R-CNN+LSS+DCN" by applying the local sampling strategy and applying one deformable convolution layer to the query frame. As a result, by comparing ``MaskTrack R-CNN+LSS+DCN" and our method, we could obtain the performance gain purely brought by temporal feature calibration.}

As shown in Table~\ref{error_analysis}, \myTextColor{the three types of error rates all significantly increase when the occlusion level increases. Among them, the segmentation error rate increases the most, from 12.1\% to 34.1\% for MaskTrack R-CNN}, which demonstrates that severe occlusion will greatly increase the difficulty of the segmentation task. In this sense, accurately localizing the object is helpful for mitigating the impact of occlusions.
\myTextColor{Meanwhile, among the three error types, the error rate of classification is much higher than that of segmentation and tracking. So a better classification result is important to improving the overall performance.}

\myTextColor{One could also observe that (1) no matter in terms of classification error rate, segmentation error rate, or ID switch rate, the gain of our method over ``MaskTrack R-CNN+LSS+DCN" increases when the occlusion level increases (\eg, CMaskTrack R-CNN decreases the classification error rate by 2.7\%, 3.9\%, and 6.6\% respectively); (2) in terms of segmentation error rate and ID switch rate, the gain of ``MaskTrack R-CNN+LSS+DCN" over the baseline ``MaskTrack R-CNN" does not change too much when the occlusion level increases (\eg, ``MaskTrack R-CNN+LSS+DCN" decreases the segmentation error rate by 6.1\%, 6.4\%, and 6.5\% respectively); (3) in terms of classification error rate, the gain of ``MaskTrack R-CNN+LSS+DCN" over the baseline ``MaskTrack R-CNN" even decreases when the occlusion level increases (No occlusion: 11.1\%, Slight occlusion: 8.1\%, and Severe occlusion: 4.3\%).}

\myTextColor{By comparing Observation (1), (2), and (3), one could conclude that the TFC module improves more in occluded scenes compared with using other training strategies (\eg, the local sampling strategy) and model structure (\eg, applying deformable convolution to the query frame). The same conclusion is also drawn if we compare the relative error decreasing rate.}

\paragraph{\myTextColor{Effect of Better Feature Representations.}}~\myTextColor{To test the effect of better feature representations on occlusion, we further try Swin-T~\cite{swin} and ResNeXt-101~\cite{resnext} backbone on MaskTrack R-CNN and QueryVIS. As can be seen in Table~\ref{tb:backbone}, both Swin-T and ResNeXt-101 achieve great improvement (about 4 AP) on OVIS. And these larger backbones can also achieve obvious AP improvement at all occlusion levels.}

\begin{table*}
\centering
\begin{tabular}{|l|c|p{0.48cm}<{\centering}p{0.48cm}<{\centering}p{0.57cm}<{\centering}p{0.48cm}<{\centering}p{0.6cm}<{\centering}|p{0.6cm}<{\centering}p{0.6cm}<{\centering}p{0.6cm}<{\centering}|}
\hline
Methods & Backbone & AP  & $\text{AP}_{50}$  & $\text{AP}_{75}$  & $\text{AR}_{1}$  & $\text{AR}_{10}$ & $\text{AP}_{SO}$  & $\text{AP}_{MO}$  & $\text{AP}_{HO}$ \\
\hline
\hline
\multirow{3}{*}{MaskTrack R-CNN~\cite{youtube_vis}} & ResNet-50 & 10.8 & 25.3 & 8.5 & 7.9 & 14.9 & 23.0 & 12.8 & 2.7 \\
& Swin-T & 14.0 & 30.5 & 11.5 & 9.2 & 19.2 & 26.9 & 16.0 & 3.7 \\
& ResNeXt-101 & \textbf{14.6} & \textbf{33.5} & \textbf{12.0} & \textbf{9.6} & \textbf{19.4} & \textbf{27.1} & \textbf{16.7} & \textbf{3.8} \\
\hline
\hline
\multirow{3}{*}{QueryVIS~\cite{queryinst}} & ResNet-50 & 12.8 & 28.8 & 11.0 & 8.6 & 19.2 & 25.2 & 15.1 & 2.6 \\
& Swin-T & 16.5 & 36.2 & 14.5 & 10.2 & 22.6 & 31.2 & \textbf{19.4} & 4.2 \\
& ResNeXt-101 & \textbf{16.9} & \textbf{36.5} & \textbf{14.7} & \textbf{10.6} & \textbf{23.5} & \textbf{31.8} & 19.2 & \textbf{4.6} \\
\hline
\end{tabular}
\caption{\myTextColor{Effect of larger backbones. For a fair comparison, the results shown here are all trained only 12 epochs for both pre-training on COCO and training on OVIS.}}
\label{tb:backbone}
\end{table*}

\paragraph{\myTextColor{Effect of Larger Input Resolutions.}}~\myTextColor{We try to replace the 640$\times$360 input resolution with 1280$\times$720 which is similar to the commonly used input resolution for COCO~\cite{coco}. As shown in Table~\ref{tb:resolution}, when the input resolution increases, the performance improves slightly (0.5 AP for MaskTrack R-CNN~\cite{youtube_vis} and 0.3 AP for SipMask~\cite{sipmask}).}

\begin{table*}
\centering
\begin{tabular}{|l|c|p{0.48cm}<{\centering}p{0.48cm}<{\centering}p{0.57cm}<{\centering}p{0.48cm}<{\centering}p{0.6cm}<{\centering}|p{0.6cm}<{\centering}p{0.6cm}<{\centering}p{0.6cm}<{\centering}|}
\hline
Methods & Input size & AP  & $\text{AP}_{50}$  & $\text{AP}_{75}$  & $\text{AR}_{1}$  & $\text{AR}_{10}$ & $\text{AP}_{SO}$  & $\text{AP}_{MO}$  & $\text{AP}_{HO}$ \\
\hline
\hline
\multirow{2}{*}{MaskTrack R-CNN~\cite{youtube_vis}} & 640$\times$360 & 10.8 & 25.3 & 8.5 & \textbf{7.9} & 14.9 & 23.0 & 12.8 & \textbf{2.7} \\
& 1280$\times$720 & \textbf{11.3} & \textbf{25.8} & \textbf{9.3} & \textbf{7.9} & \textbf{15.7} & \textbf{23.9} & \textbf{12.9} & 2.6 \\
\hline
\hline
\multirow{2}{*}{SipMask~\cite{sipmask}} & 640$\times$360 & 10.2 & \textbf{24.7} & 7.8 & \textbf{7.9} & \textbf{15.8} & \textbf{19.9} & 10.5 & \textbf{2.2} \\
& 1280$\times$720 & \textbf{10.5} & 24.3 & \textbf{8.4} & 7.1 & 15.7 & \textbf{19.9} & \textbf{11.8} & 2.0 \\
\hline
\end{tabular}
\caption{\myTextColor{Effect of larger input resolutions.}}
\label{tb:resolution}
\end{table*}

\paragraph{\myTextColor{Methods Specifically Designed for Occlusion.}}~\myTextColor{We also migrate three image-level detection methods to the CMaskTrack R-CNN model, including 1) the repulsion loss~\cite{repulsion} which requires the predicted boxes to keep away from other ground-truth boxes; 2) the compact loss~\cite{orcnn} which enforces proposals to be close and locate compactly to the corresponding ground-truth; 3) the occluder branch~\cite{bcnet} (without any extra designs like the Non-local~\cite{nonlocal} operation and boundary prediction) which additionally learns the feature of occluders with a new branch and then fuses the feature of occluders and occludees. In particular, the repulsion loss and compact loss are specifically designed for crowded pedestrian detection, and the occluder branch is designed for the occlusion problem of common objects.}

\myTextColor{As shown in Table~\ref{tb:occlusion_methods}, the compact loss and occluder branch improve $\text{AP}_{HO}$ by 0.4 and 0.3 respectively, while their overall AP improvements are marginal. We believe more gains can be achieved by developing more delicate occlusion handling algorithms and leveraging occluded data (see Sec.~\ref{sec:future_directions} for future work discussion).}

\begin{table*}
\centering
\begin{tabular}{|l|p{0.48cm}<{\centering}p{0.48cm}<{\centering}p{0.57cm}<{\centering}p{0.48cm}<{\centering}p{0.6cm}<{\centering}|p{0.6cm}<{\centering}p{0.6cm}<{\centering}p{0.6cm}<{\centering}|}
\hline
Methods & AP  & $\text{AP}_{50}$  & $\text{AP}_{75}$  & $\text{AR}_{1}$  & $\text{AR}_{10}$ & $\text{AP}_{SO}$  & $\text{AP}_{MO}$  & $\text{AP}_{HO}$ \\
\hline
\hline
CMaskTrack R-CNN & 15.4 & 33.9 & 13.1 & 9.3 & 20.0 & \textbf{28.6} & \textbf{18.7} & 4.1 \\
$+$ Repulsion loss~\cite{repulsion} & 14.7 & 32.0 & \textbf{13.8} & 9.2 & 19.3 & 26.9 & 17.7 & 4.0 \\
$+$ Compact loss~\cite{orcnn} & 15.4 & 34.1 & 12.7 & 9.4 & 19.4 & 27.9 & 18.3 & \textbf{4.5} \\
$+$ Occluder branch~\cite{bcnet} (w/o extra designs) & \textbf{15.6} & \textbf{34.3} & 13.5 & \textbf{9.7} & \textbf{20.1} & 28.3 & 17.9 & 4.4 \\
\hline
\end{tabular}
\caption{\myTextColor{Effect of three existing occlusion handling methods that are specifically designed for image-level detection tasks. ``w/o extra designs'' means that we remove the Non-local operation~\cite{nonlocal} and boundary prediction in the occluder branch for a fair comparison with other methods.}}
\label{tb:occlusion_methods}
\end{table*}

\begin{figure}[t]
\centering
 \includegraphics[width=0.95\linewidth]{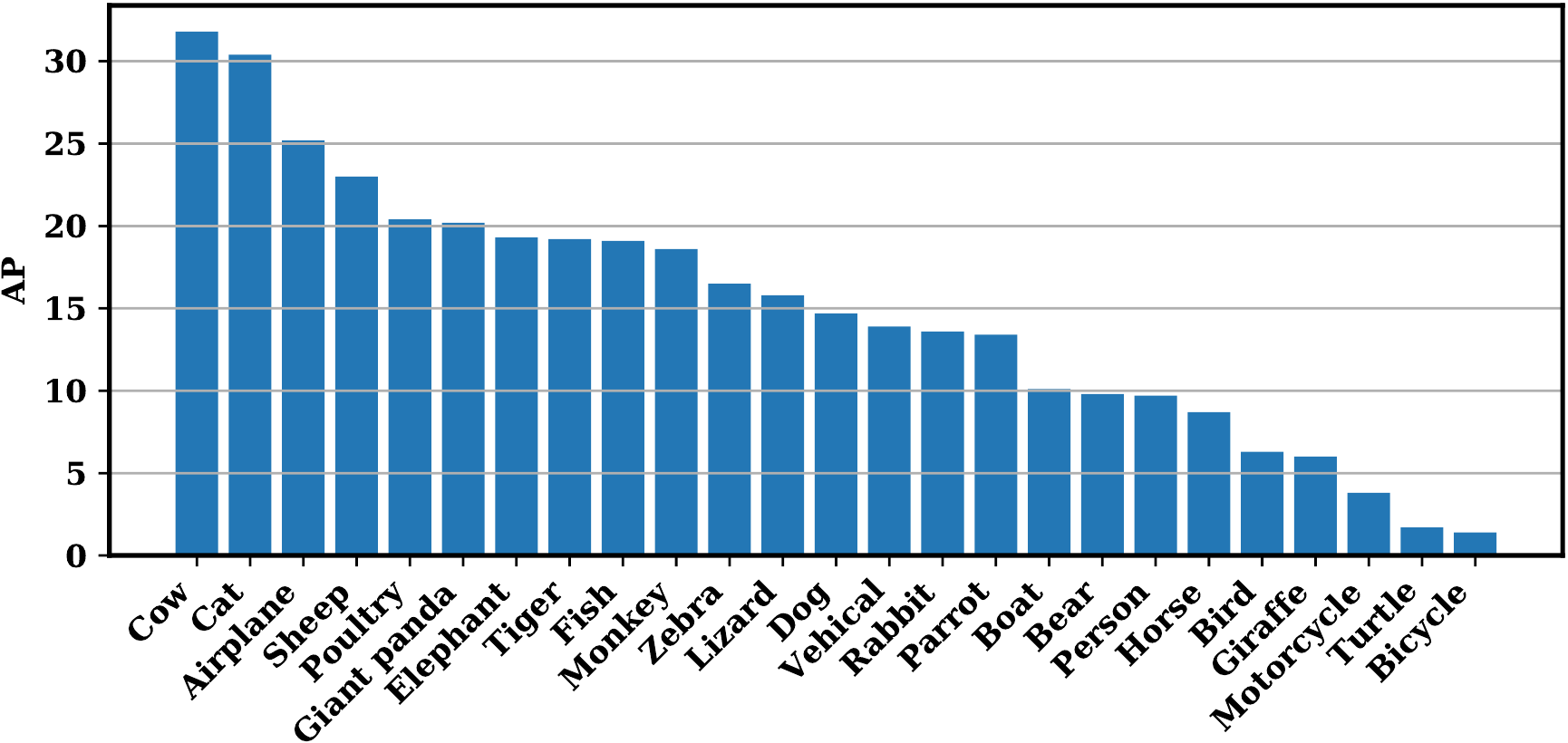}
 \caption{Per-class AP of CMaskTrack R-CNN on OVIS.}
\label{fig:ap_per_class}
\end{figure}

\begin{figure}[t]
\centering
 \includegraphics[width=0.8\linewidth]{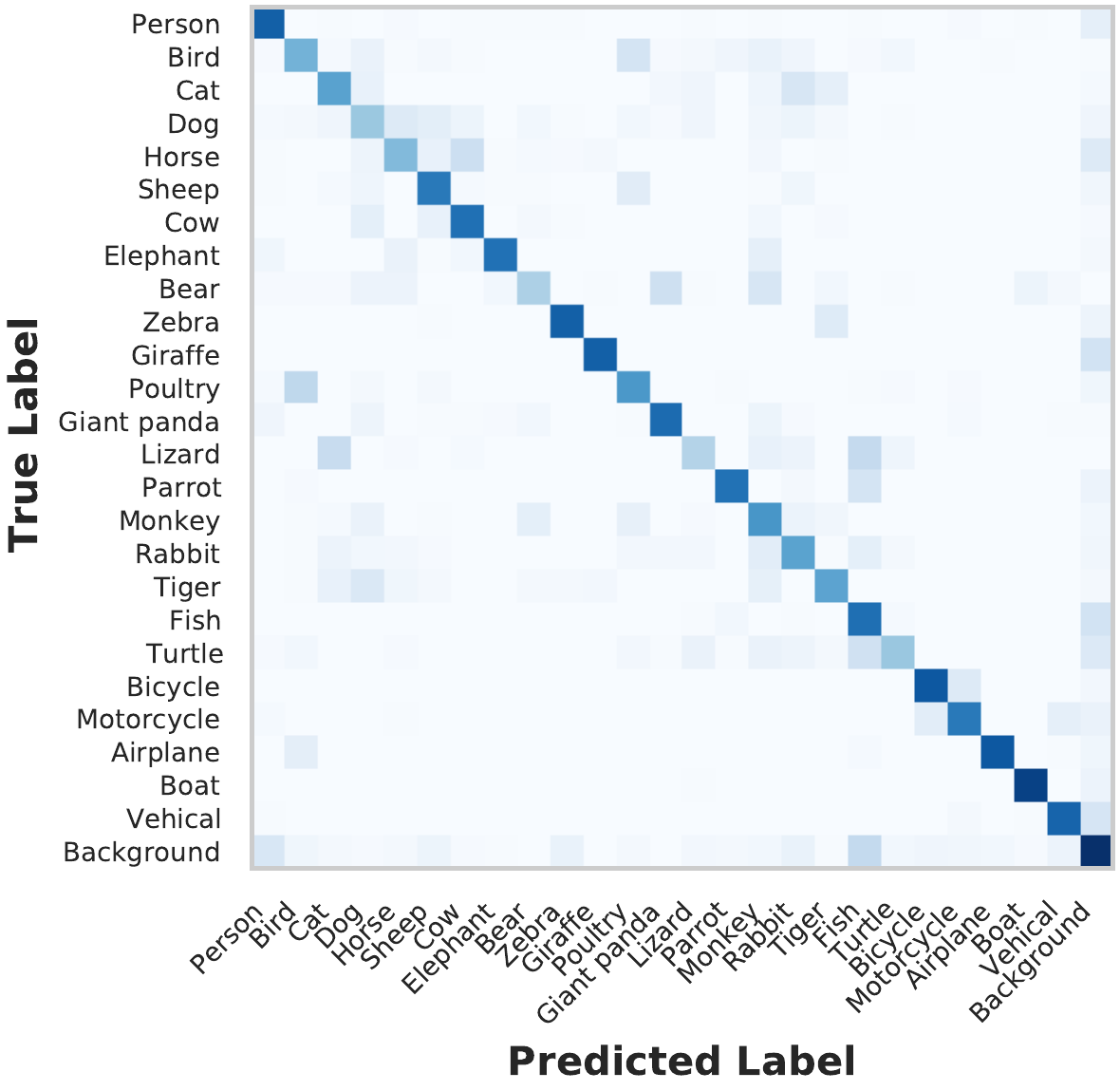}
 \caption{Confusion matrix for classification.}
\label{fig:cm}
\end{figure}

\paragraph{Per-class Results.}~The per-class AP scores of CMaskTrack R-CNN are shown in Fig.~\ref{fig:ap_per_class}. It shows that the Top-5 challenging categories are Bicycle, Turtle, Motorcycle, Giraffe, and Bird. The confusion matrix is also given in Fig.~\ref{fig:cm}. As it shows, most categories can be correctly classified except for some visually similar category pairs (\eg,~Poultry and Bird, Bicycle and Motorcycle).

\paragraph{Ablation Study of the TFC Module.}~To verify the rationality of the TFC module, we firstly test the effect of the local sampling strategy of reference frames during training. As shown in Table~\ref{tb:localsampling}, by only sampling the reference frames locally within $\epsilon_\text{train}=5$ frames instead of sampling in the whole video, \myTextColor{MaskTrack R-CNN, SipMask, and QueryVIS all obtain significant AP improvements of 2.7, 2.6, and 1.7 respectively}, which demonstrates that the local sampling strategy of reference frames during training is necessary and beneficial to learn how to track objects in the long videos of OVIS.

\begin{table*}
\centering
\begin{tabular}{|l|c|p{0.48cm}<{\centering}p{0.48cm}<{\centering}p{0.57cm}<{\centering}p{0.48cm}<{\centering}p{0.6cm}<{\centering}|p{0.6cm}<{\centering}p{0.6cm}<{\centering}p{0.6cm}<{\centering}|}
\hline
Methods & Local sampling & AP  & $\text{AP}_{50}$  & $\text{AP}_{75}$  & $\text{AR}_{1}$  & $\text{AR}_{10}$ & $\text{AP}_{SO}$  & $\text{AP}_{MO}$  & $\text{AP}_{HO}$ \\
\hline
\hline
\multirow{2}{*}{MaskTrack R-CNN~\cite{youtube_vis}}        & & 10.8 & 25.3 & 8.5 & 7.9 & 14.9 & 23.0 & 12.8 & 2.7\\
& \checkmark & \textbf{13.5} & \textbf{29.9} & \textbf{11.3} & \textbf{8.5} & \textbf{18.7} & \textbf{25.4} & \textbf{16.7} & \textbf{3.3}\\
\hline
\multirow{2}{*}{SipMask~\cite{sipmask}}        & & 10.2 & 24.7 & 7.8 & 7.9 & 15.8 & 19.9 & 10.5 & 2.2\\
& \checkmark & \textbf{12.8} & \textbf{29.8} & \textbf{9.6} & \textbf{8.7} & \textbf{17.9} & \textbf{25.5} & \textbf{14.7} & \textbf{2.5}\\
\hline
\multirow{2}{*}{QueryVIS~\cite{queryinst}}      & & 14.7 & 34.7 & 11.6 & 9.0 & 21.2 & 27.3 & 17.2 & 4.1 \\
& \checkmark & \textbf{16.4} & \textbf{37.8} & \textbf{12.4} & \textbf{9.9} & \textbf{22.9} & \textbf{31.2} & \textbf{19.0} & \textbf{4.3}\\
\hline
\end{tabular}
\caption{Effect of the local sampling strategy on the OVIS validation set.}
\label{tb:localsampling}
\end{table*}

We further study the temporal feature calibration module with a few alternatives. The first option is a naive combination, which sums up the feature of the query frame and the reference frame without any feature alignment. The second option is to replace the correlation operation in our module by calculating the element-wise difference between feature maps, which is similar to the operation used in~\cite{maskprop}.
We denote the three options as ``$+$ Uncalibrated Addition" and ``$+$ $\text{Calibration}_{\text{diff}}$" respectively and our module as ``$+$ $\text{Calibration}_{\text{corr}}$" in Table~\ref{tb:ablation}.

\begin{table*}
\centering
\begin{tabular}{|l|c|p{0.48cm}<{\centering}p{0.48cm}<{\centering}p{0.57cm}<{\centering}p{0.48cm}<{\centering}p{0.6cm}<{\centering}|p{0.6cm}<{\centering}p{0.6cm}<{\centering}p{0.6cm}<{\centering}|}
\hline
Methods & Local sampling & AP  & $\text{AP}_{50}$  & $\text{AP}_{75}$  & $\text{AR}_{1}$  & $\text{AR}_{10}$ & $\text{AP}_{SO}$  & $\text{AP}_{MO}$  & $\text{AP}_{HO}$ \\
\hline
\hline
MaskTrack R-CNN~\cite{youtube_vis}       & & 10.8 & 25.3 & 8.5 & 7.9 & 14.9 & 23.0 & 12.8 & 2.7\\ 
MaskTrack R-CNN~\cite{youtube_vis}& \checkmark & 13.5 & 29.9 & 11.3 & 8.5 & 18.7 & 25.4 & 16.7 & 3.3 \\
\cite{youtube_vis} $+$ DCN & \checkmark & 14.0 & 31.2 & 11.2 & 8.8 & 18.5 & 26.2 & 16.2 & 3.2\\
\hline
\hline
\cite{youtube_vis} $+$ Uncalibrated Addition & \checkmark & 12.9 & 29.4 & 11.5 & 8.2 & 16.6 & 25.6 & 15.3 & 3.1\\ 
\cite{youtube_vis} $+$ $\text{Calibration}_{\text{diff}}$ & \checkmark & 14.4 & 32.6 & 12.3 & 8.6 & 18.9 & 25.3 & 17.6 & 3.8\\
\cite{youtube_vis} $+$ $\text{Calibration}_{\text{corr}}$ & \checkmark & \textbf{15.4} & \textbf{33.9} &  \textbf{13.1} & \textbf{9.3} & \textbf{20.0} &\textbf{28.6} & \textbf{18.7} & \textbf{4.1}\\
\hline
\end{tabular}
\caption{Effect of the local sampling strategy and the comparison of different feature fusion methods. ``Local sampling" means only sample the reference frames locally within $\epsilon_\text{train}=5$ frames during training. \myTextColor{``$+$ DCN'' means applying a deformable convolutional layer on the query frame itself.} ``$+$ Uncalibrated Addition" means adding feature maps directly without calibration. ``$+$ $\text{Calibration}_{\text{diff}}$" means generating the calibration offset based on the element-wise difference between feature maps, similar to \cite{maskprop} did. ``$+$ $\text{Calibration}_{\text{corr}}$" is the presented method in Sec.~\ref{sec:implementation}.}
\label{tb:ablation}
\end{table*}

As we can see, with the enhanced MaskTrack R-CNN (with local sampling strategy of reference frames during training) as the base model, the naive ``$+$ Uncalibrated Addition" combination even degrades the final AP. This is because the direct addition of the uncalibrated features from other frames may bring noises to the object localization process. In contrast, after applying feature calibration, the performance is improved. ``$+$ $\text{Calibration}_{\text{corr}}$" achieves an AP of 15.4, an improvement of 1.9 over the baseline method without feature fusion and 1.0 over ``$+$ $\text{Calibration}_{\text{diff}}$". We argue that the correlation operation is able to provide a richer context for feature calibration because it calculates the similarity between the query position and its neighboring positions. \myTextColor{Testing on P-100 GPU, the speed of CMaskTrack R-CNN when using $\text{Calibration}_{\text{diff}}$ and $\text{Calibration}_{\text{corr}}$ are 16 and 7 fps respectively.}

\begin{figure}[t]
\centering
  \includegraphics[width=0.95\linewidth]{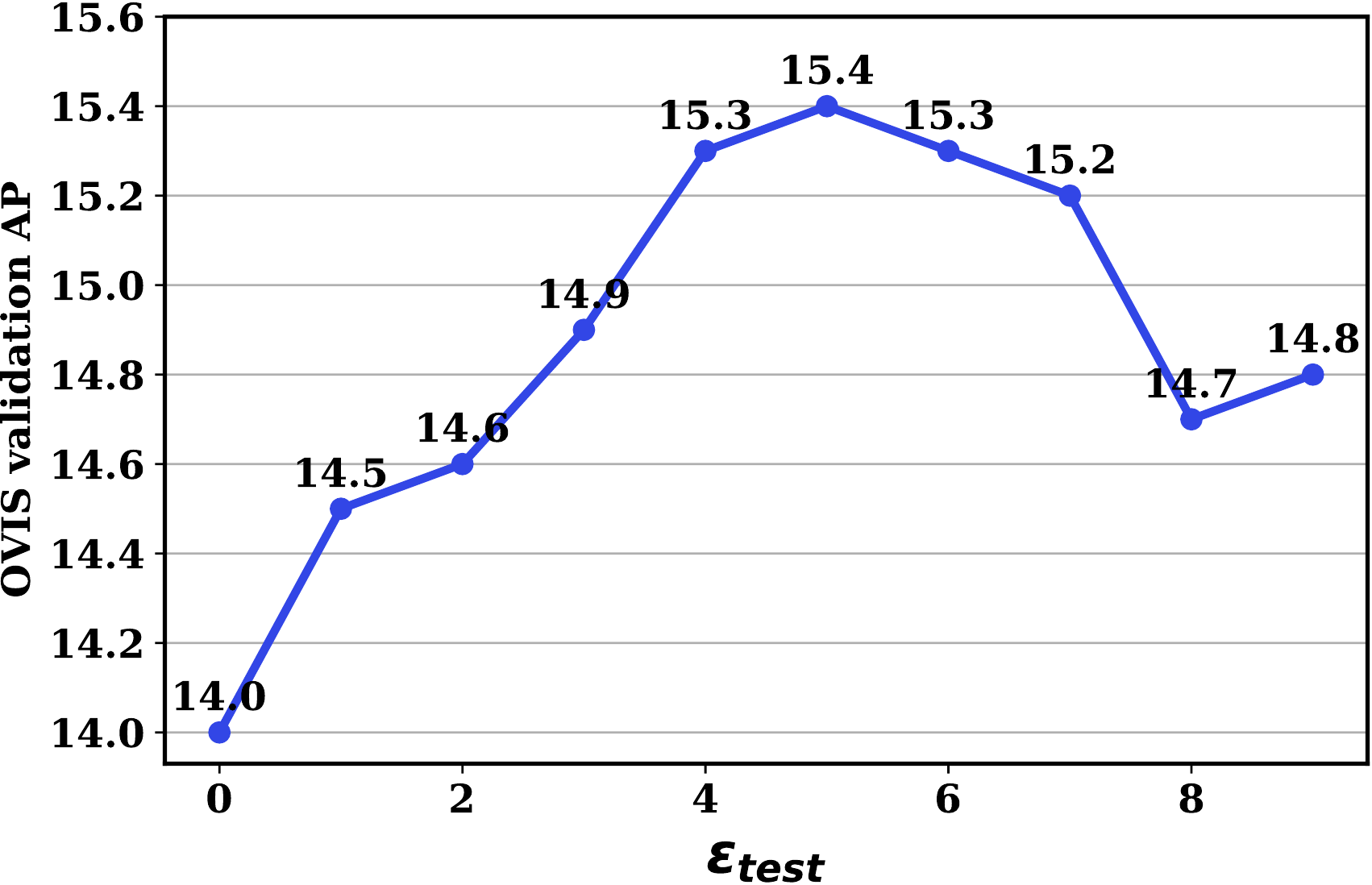}
  \caption{Results of different reference frame range $\epsilon_{test}$ on the OVIS validation set. Notably, $\epsilon_{test}=0$ indicates applying the deformable convolutional layer to the query frame itself, without leveraging adjacent frames.}
\label{fig:epsilon}
\end{figure}

We also conduct experiments to analyze the influence of the reference frames range $\epsilon_{test}$. $\epsilon_{test}=0$ means applying the deformable convolutional layer to the query frame itself. As can be seen in Fig.~\ref{fig:epsilon}, the AP increases when $\epsilon_{test}$ increases, and reaching the highest value at $\epsilon_{test}=5$. Even if $\epsilon_{test}=1$, the performance exceeds the setting of $\epsilon_{test}=0$, which demonstrates that calibrating features from adjacent frames is beneficial to video instance segmentation.

\begin{figure}
\centering
\includegraphics[width=0.95\linewidth]{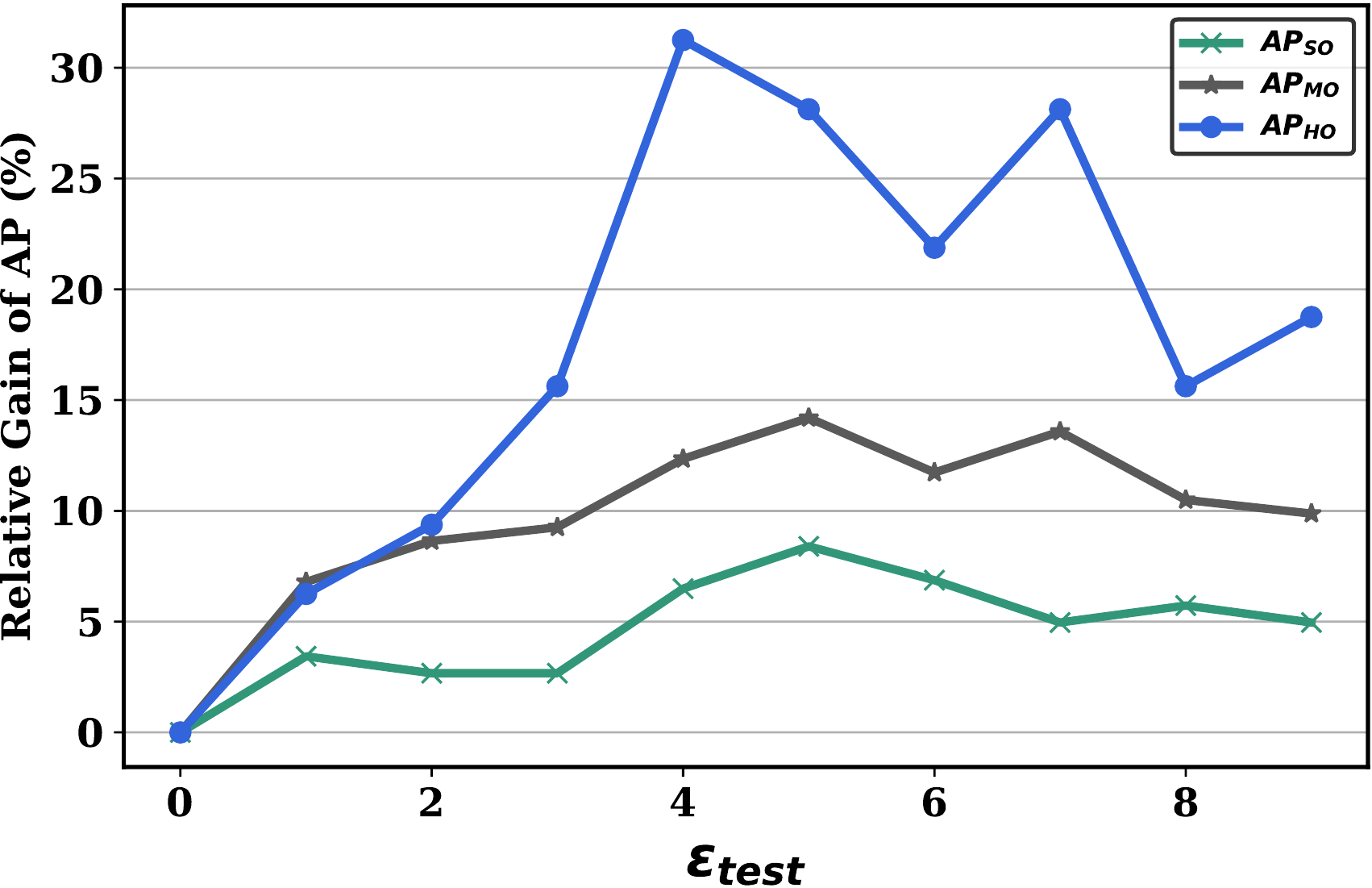}
\caption{\myTextColor{Relative gain of different occlusion levels with increasing reference frame range $\epsilon_{test}$.}}
\label{fig:epsilon_test_occlusion_level}
\end{figure}

\myTextColor{To further compare the improvement of TFC on different occlusion levels, we evaluate the relative gain of AP on different occlusion levels for a fair comparison. As shown in Fig.~\ref{fig:epsilon_test_occlusion_level}, we report the relative gain by varying $\epsilon_{test}$. The larger the $\epsilon_{test}$ is, the more temporal context will be aggregated. As can be seen, the relative gain of $\text{AP}_{HO}$ is much higher than that of $\text{AP}_{MO}$. The relative gain of $\text{AP}_{SO}$ is smallest once the temporal context is considered ($\epsilon_{test}\textgreater 0$). The result demonstrates the effectiveness of temporal feature aggregation on occlusion handling.}

\section{\myTextColor{Future Directions}}
\label{sec:future_directions}

\myTextColor{In the future, there are still many interesting issues that can be studied and many remaining difficulties to be addressed with OVIS, such as:}

\paragraph{\myTextColor{Occlusion-aware models.}}~\myTextColor{Effectively handling occlusions is one of the most straightforward ways to improve the performance in OVIS. In terms of occlusion-aware models, there are a few directions that can be exploited in our future work. For example, compositional models~\cite{compositional,compositional2,compositional3} might be a good choice as they are robust to partial occlusions. It is also interesting to test if completing the invisible parts of occluded objects (\emph{a.k.a.}~de-occlusion~\cite{deocclusion}) is useful in this scenario.}

\paragraph{\myTextColor{Occluded data generation.}}~\myTextColor{Due to the high cost of annotation, the scale of video instance segmentation datasets is relatively smaller than image datasets. Some works~\cite{cutout,cutmix,copypaste_det,copypaste_instseg} have proposed augmenting the common datasets (\eg, COCO~\cite{coco}) with partial occlusions, and some works~\cite{nikolenko2019synthetic,metasim,metasim2} synthesize structured amodal data in occluded scenes using simulators. It can be anticipated that utilizing those data with proper training paradigms will improve the performance in VIS.}

\paragraph{\myTextColor{Learning from occlusion annotations.}}~\myTextColor{In OVIS, a coarse annotation of occlusion levels (no occlusion, slight occlusion, and server occlusion) is given per object. As a prior knowledge that can be accessed during training, learning paradigms that can abstract such information deserve special attention.}

\paragraph{\myTextColor{Large scale model pre-training.}}~\myTextColor{According to our experiments, it improves the performance to conduct joint training with image datasets. With the development of self-supervised learning~\cite{mae}, exploiting the unlimited amounts of unlabeled data for model pre-training, then transferring the pre-trained model into OVIS will largely enhance the discriminative power of frame embeddings.}

\paragraph{\myTextColor{Dataset versatility.}}~\myTextColor{At last, we are also interested in formalizing the experimental track of OVIS for video object segmentation, either in an unsupervised, semi-supervised, or interactive setting. It is also of paramount importance to extend OVIS to video panoptic segmentation~\cite{kim2020video}. We believe the OVIS dataset will trigger more research in understanding videos in complex and diverse scenes.}

\section{Conclusions}
In this work, we target video instance segmentation in occluded scenes and accordingly contribute a large-scale dataset called OVIS. OVIS consists of 296k high-quality instance masks of 5,223 heavily occluded instances. While being the second benchmark dataset after YouTube-VIS, OVIS is designed to examine the ability of current video understanding systems in terms of handling object occlusions. A general conclusion is that the baseline performance on OVIS is far below that on YouTube-VIS, which suggests that more effort should be devoted in the future to tackling object occlusions or de-occluding objects~\cite{deocclusion}. We also explore ways about leveraging temporal context cues to alleviate the occlusion matter and conduct a comprehensive analysis of occlusion handling on OVIS.

\begin{acknowledgements}
This work is supported by Turing AI Fellowship EP/W002981/1.
\end{acknowledgements}

%
%

\bibliographystyle{spmpsci}      
\bibliography{ijcv.bib}   


\end{sloppypar}
\end{document}